\documentclass[10pt,letterpaper]{article}

\usepackage[ruled,vlined,linesnumbered]{algorithm2e}
\usepackage{amsmath}
\usepackage{amssymb}
\usepackage[toc,page]{appendix}
\usepackage{array}
\usepackage{bm}
\usepackage{booktabs}
\usepackage[wby]{callouts}
\usepackage[aboveskip=1pt,labelfont=bf,labelsep=period,justification=raggedright,singlelinecheck=off]{caption}
\usepackage{changepage}
\usepackage{cite}
\usepackage{enumitem}
\usepackage{epstopdf}
\usepackage{fancyhdr}
\usepackage{float}
\usepackage[T1]{fontenc}
\usepackage[top=0.9in,left=1in,footskip=0.75in]{geometry}
\usepackage{graphicx}
\usepackage[hidelinks]{hyperref}
\usepackage[utf8]{inputenc}
\usepackage{lastpage}
\usepackage[right]{lineno}
\usepackage{listings}
\usepackage{marvosym}
\usepackage[nopatch=eqnum]{microtype}
\usepackage{multicol}
\usepackage{multirow}
\usepackage{nameref}
\usepackage{parskip}
\usepackage{textcomp}
\usepackage{tikz}
\usepackage{titlesec}
\usepackage{utfsym}
\usepackage{varwidth}
\usepackage[table,dvipsnames]{xcolor} \usepackage[noabbrev,nameinlink,capitalize,poorman]{cleveref}
\usepackage{tabularx}

\DisableLigatures[f]{encoding = *, family = * }

\raggedright
\makeatletter
\renewcommand{\@biblabel}[1]{\quad#1.}
\makeatother

\fancyheadoffset[L]{2.25in}
\fancyfootoffset[L]{2.25in}
\graphicspath{{./Pictures}}

\newlist{enumsymbol}{description}{1}
\setlist[enumsymbol,1]{
	labelwidth=25pt,
	itemindent=10pt,
	itemsep=-3pt,
	topsep=0pt,
}

\newcommand{\graybox}[1]{\begin{center}\colorbox{Gray!5}{\begin{varwidth}{.85\linewidth}\hspace{1ex}#1\hspace{1ex}\end{varwidth}}\end{center}}
	
\newcommand{\smallgraybox}[1]{\begin{center}\colorbox{Gray!5}{\begin{varwidth}{.95\linewidth}#1\end{varwidth}}\end{center}}
\newcommand{\tcaret}{\textasciicircum{}}
\newcommand{\tminus}{\textendash{}}

\newcommand{\eks}{\textsf{x}}

\tikzset{
  callout font/.style={font=\sffamily\tiny}
}

\lstdefinestyle{py}{
  language=Python,
  basicstyle=\ttfamily\footnotesize,
  columns=fullflexible,
  keepspaces=true,
  showstringspaces=false,
  commentstyle=\itshape,
  keywordstyle=\ttfamily\bfseries,
}

\lstnewenvironment{plisting}[1][]
    {\nolinenumbers\lstset{#1}}
    {\linenumbers}

\crefname{figure}{Fig}{Figs}
\Crefname{figure}{Fig}{Figs}

\crefname{equation}{Eq}{Eqs}
\Crefname{equation}{Eq}{Eqs}

\crefname{table}{Table}{Tables}
\Crefname{table}{Table}{Tables}

\crefname{section}{section}{sections}
\Crefname{section}{Section}{Sections}

\begin{document}
\vspace*{0.2in}

\begin{flushleft}
{\Large
\textbf{PALMS: A Computational Implementation for Pavlovian Associative Learning Models' Simulation}
}
\newline
\\
Martin~Fixman\textsuperscript{1,2},
Alessandro~Abati\textsuperscript{1,2},
Julián~Jiménez~Nimmo\textsuperscript{1,2},
Sean~Lim\textsuperscript{1,2},
Esther~Mondragón\textsuperscript{1,2*}
\\
\bigskip
\textbf{1} Artificial Intelligence Research Centre (CitAI), Department of Computer Science, City St George's, University of London, London, United Kingdom
\\
\textbf{2} Centre for Computational and Animal Learning Research, CAL-R, \url{https://cal-r.org/}
\bigskip

* \href{mailto:e.mondragon@citystgeorges.ac.uk}{e.mondragon@citystgeorges.ac.uk}

\end{flushleft}

\section*{Abstract}
In contrast to static formalisms, computational definitions describe the operational mechanisms of a model. Simulations are an essential part of the cycle of theory development and refinement, assisting researchers in formulating the precise definitions that models require, and making accurate predictions. This manuscript introduces a computational implementation of Pavlovian learning models in a Python environment, termed Pavlovian Associative Learning Models' Simulation (PALMS). In addition to the canonical Rescorla-Wagner model, attentional approaches are implemented, including Pearce-Kaye-Hall, Mackintosh Extended, Le Pelley's Hybrid, and a novel extension of the Rescorla-Wagner model featuring a unified variable learning rate that synthesises Mackintosh's and Pearce and Hall's opposing conceptualisations. To our knowledge, only the first attentional model has been previously specified computationally in a general design tool. PALMS integrates a graphical interface that permits the input of entire experimental designs in an alphanumeric format, akin to that used by experimental neuroscientists. It uniquely enables the simulation of experiments involving hundreds of stimuli, such as those used with human participants, and the computation of configural cues and configural-cue compounds across all models, thereby substantially broadening their predictive capabilities. A comprehensive description of the models' implementation and the environment functionalities is provided in the paper; these include efficient and accurate operation and instant visualisation of predicted results across different models within a single architecture and environment. We evaluate PALMS by simulating five published experiments in the associative learning literature that assessed the predictive scope of existing models, and we show that this implementation provides neuroscientists with a useful tool for identifying critical variables, refining experimental designs, making precise predictions, comparing model fitness, and formulating new theoretical approaches.
PALMS is licensed under the open-source GNU Lesser General Public License 3.0. The environment source code and the latest multiplatform release build are accessible as a GitHub repository at \url{https://github.com/cal-r/PALMS-Simulator}.

\section*{Author summary}
Research on associative learning is multidisciplinary, encompassing disciplines such as neuroscience, AI, psychology, psychiatry, behavioural sciences, planning, and marketing. Unlike static formalisms, precise computational definitions specify how a model operates, enabling model simulation, swift and error-free prediction calculations, which are essential for testing theories, comparing predictions, holding models accountable, and providing a common language across fields.  We introduce Pavlovian Associative Learning Models' Simulation (PALMS), a user-friendly, open-source Python environment for simulating classical conditioning and studying the role of attention in learning. PALMS implements the prescriptive Rescorla-Wagner and attentional models: Pearce-Kaye-Hall, Mackintosh Extended, Le Pelley's Hybrid, and a new hybrid model with a unified variable learning rate that blends Mackintosh and Pearce-Hall's conflicting views. Its graphical interface makes it easy for neuroscientists to enter experiments. Our computational implementation supports simulations with hundreds of stimuli, configural cues, and compounds, broadening the models' predictive power. Designed for efficiency, it offers instant visual results and useful features. We evaluate PALMS by simulating five published experiments, highlighting its value for model comparison and refinement, and, more generally, as a tool to assist research.


\section{Introduction}

In contrast to static mathematical formalisations, computational implementations describe the operational mechanism of a model, explicitly outlining the interconnected stages that can be expressed in detailed algorithms, thereby enabling researchers to conduct accurate simulations. Simulations are essential for scientific development. Firstly, they facilitate the formulation of the precise definitions necessary to implement a model, providing the means to test the model's predictions and hold it accountable. Secondly, they enable swift and accurate outcome calculations \cite{alonso2014have}.

In this paper, we present computational specifications for five associative learning models and implement them in a Python environment that provides a comprehensive foundation for other researchers to define and test their own models, empowering them to efficiently code, explore new formulations, and evaluate results directly and quickly.
This work also contributes considerably to the so-called Three Rs initiatives, as alternatives to animals in research, namely, replacement, reduction, and refinement \cite{neuhaus2022rise}.

The formation of associations between repeatedly paired events is a fundamental principle in learning theory. In Pavlovian conditioning, the association of two stimuli enables individuals to anticipate the occurrence of an event, the outcome, from the presentation of its correlated cue, or predictor \cite{rescorla_1988_pavlovian}.

Associative learning processes have long been assumed to underlie other more complex cognitive phenomena \cite{shanks_2010}.
For instance, they are considered to be at the basis of human causality and categorisation judgment \cite{benton2024associative,hansel2024contiguity}.
Learning theory also plays a crucial role in the development of computational psychiatry and clinical studies \cite{fineberg162associative,mondragon2024mediated,sapey2022associative} and neuroscience \cite{avvisati2024distributional,held2024associative}.
It has also contributed to areas as disparate as planning \cite{E_Lind_2018}, marketing \cite{E_Elsen_2025} and even molecular chemistry \cite{E_Molecules}, among others.
Critically, associative learning has inspired, and continues to do so, other theoretical frameworks in cognitive science and artificial intelligence.
For example, the idea of minimising reward prediction has been incorporated in artificial computational models of learning \cite{anokhin2024associative,sosa2024conditioned}.

Following early-20th-century attempts to describe the learning processes mathematically \cite{Bush1951_BUSAMM,Estes50}, a substantial number of computational models have been developed to capture associative learning mechanisms, with varying assumptions and complexity.

One such model, the Rescorla-Wagner model (RW, henceforth) \cite{RescorlaWagner}, a breakthrough over previous linear operator models \cite{Bush1951_BUSAMM,Hull_1943}, is considered to be a landmark in the development of learning theory \cite{Esber2025,miller1995assessment,navarro2019between,wasserman2022}. Since its publication, this model has consistently ranked among the top 10 most-cited papers in experimental psychology \cite{soto2023rescorla}.

RW's simple and powerful approach to prediction relies on the notion of global error correction and has become the obligatory reference and baseline comparison for other models of Pavlovian conditioning \cite{gershman2012exploring,kang2024bayesian,soto2023rescorla,kokkola2019double}, setting the standards for associative theory \cite{Hall1991,Roiblat87}.
Thus, many newly developed models have been outlined that preserve Rescorla-Wagner's core assumptions while seeking to accommodate the phenomena that lie outside its scope. Among these, the so-called attentional models challenged the perhaps over-simplistic RW's assumption of a constant learning rate, as a learning modulator solely determined by the stimulus' physical salience \cite{E_Jeong_2023,E_LePelley_Attention,E_Pelley_Two_One,E_Brain_Behavior,E_PH_elaboration}.

Since the mid-eighties, two seemingly contradictory attentional approaches have rivalled to dominate the field and describe the associative mechanism of learnt attention: Mackintosh's \cite{mackintosh1975theory} and Pearce and Hall's models \cite{pearce1980model} (later improved by Pearce, Kaye \& Hall \cite{pearce1982predictive}). Both approaches posit that the speed of learning between a predictor and an outcome, or learning rate, is proportional to the attention generated by the predictor cue at any given learning episode. However, while the former postulates that the best available predictors drive attention, the latter sustains that stimuli with unexpected or inconsistent consequences capture attention.

Thus, these models generate conflicting predictions in many scenarios: Attention à la Mackintosh predicts faster learning for cues established as good predictors, whereas, according to Pearce and Hall's model, stimuli with uncertain consequences will condition faster. Given that there is empirical support to back each divergent postulate (see \cite{pearce2010two,E_LePelley_Attention} for a review), it is crucial to possess the means to readily and accurately evaluate their predictions in genuine learning scenarios.

To the best of our knowledge, there is no unified computational platform that enables systematic comparison of these models (but see \cite{calmr_app} for an R implementation of Rescorla Wagner, Mackintosh and Pearce, Kaye and Hall). Moreover, most formalisms have not been algorithmically specified. Thus, computationally and rigorously testing models' predictions remains a challenge in the field. 

In this paper, we introduce an open-source platform that implements Pavlovian learning models in a single, accurate and reliable environment, with an integrated design that provides a fast, consolidated display of results, enabling direct model and group comparisons. Although some isolated model implementations do exist, they are not integrated within a single environment. This implementation, Pavlovian Associative Learning Models' Simulation (PALMS), also uniquely defines the computation of a large number of stimuli and of configural cues, thus considerably improving on previous published tools\cite{mondragon2013extension,WRSimPaper,RW_simulator,RW_simPLUS,PH_simulator,matlab1,turbopascal}.

PALMS implements five different associative models: Rescorla~Wagner \cite{RescorlaWagner}, Pearce~Kaye~Hall \cite{pearce1982predictive}, Mackintosh's Extended \cite{lepelley2004}, Le~Pelley's Hybrid Attentional model \cite{lepelley2004}, and a modification of Rescorla~Wagner developed in our lab, the MLAB model, which conceptualises a unified attentional rate parameter capable of replicating some of the most paradigmatic predictions of Pearce~Kaye~Hall \cite{pearce1982predictive} and Mackintosh's models \cite{mackintosh1975theory}.  To the best of our knowledge, only the Rescorla~Wagner \cite{RescorlaWagner} and Pearce~Kaye~Hall \cite{pearce1982predictive} models have been previously specified and implemented within a general design tool, capable of simulating different experimental settings by simply entering their design, without the need to modify the internal code.  

In summary, the implementation of PALMS defines a large number of stimuli that go well beyond the standard 26 letters of the English alphabet, and are necessary to simulate current human experiments (e.g.,\cite{broadhurst2005discrimination,george2024absence,novelty_mismatch_haselgrove}). It enables sequential or random trial distribution, averaging values across multiple random series. It computes compound cue values and, optionally, defines a configural cue emergent, which will undergo learning and whose values will be included in the computation of the compound. Stimulus and configural cue parameters can be set individually. 

PALMS source code and application releases are publicly available on GitHub at the following URL address \url{https://github.com/cal-r/PALMS-Simulator}. It is implemented in platform-agnostic Python, and licensed under the GNU Lesser General Public Licence \cite{lgpl}.
The environment is also provided as self-contained binaries compatible with Linux, macOS, and Windows.
The code is clean, well-commented, easy to read and work with, and the PALMS team welcomes forks and pull requests from cognitive scientists and anyone interested in the program to improve the simulator.
 
\section{Methods}
\subsection{Models}
\label{adaptive_types}
\newcommand{\VNet}{\mathbb{V}}

The implemented models are the following:

\subsubsection{Rescorla Wagner model}
\label{rescorla_wagner_at}
The Rescorla-Wagner model \cite{RescorlaWagner} is a formal error correction model that estimates how strongly a stimulus, a CS, predicts a given outcome (or US). The model posits a theoretical value, associative strength, $V$, which represents the amount of knowledge or prediction estimation that a CS holds about the US.

According to the model, learning is proportional to the discrepancy between the current outcome prediction and the putative maximal amount of strength the outcome can bear ($\lambda$) when present on a given trial, zero otherwise. Crucial in this model is the introduction of a global error term, an innovation upon other linear operator error terms such as Bush-Mosteller's (1951)\cite{bush1955mstochastic} and Hull's (1943) \cite{Hull_1943}, which takes into account all present predictors' values $V_i$ rather than only the target CS predictor's value ($V_\eks$) to update learning. Thus, learning is driven by the disparity between the experienced outcome and the accumulated prediction of all present CSs. During early CS-US training, a large prediction error results in substantial increases in associative strength. As the number of pairings increases, this error is progressively reduced until it no longer supports learning. In other words, as the strength of the prediction grows, learning shrinks because the mismatch between the outcome prediction and the outcome experienced is reduced.

The change in the associative strength $\Delta V_{\eks}$ between a CS, $\eks$, and a US is described as \cref{rw_deltaV}.

\begin{equation}
	\text{If } \kappa_{\eks}^{n+1}>0,\text{ } \Delta V_{\eks}^{n+1} =\alpha_\eks \cdot \beta \cdot \left( \lambda^n - \textstyle\sum_i V_i^{n}\cdot\kappa_{i}^{n+1} \right)
	\label{rw_deltaV}
\end{equation}
where $\kappa_{i}^{n+1}$ denotes the presence value of stimulus $i$ on trial $n+1$: 1 if present, 0 if absent. The same nomenclature will be used for all models henceforth.

The constants of proportionality of the model, $\alpha_\eks$ and $\beta$, which determine the learning rate, represent the effective salience of the CS and US, respectively, which, in RW, is given by the intensity of the stimuli. In the RW model, the learning rate is thus invariant and determined by the stimulus salience.

Two distinct $\beta$ values, for a present, $\beta^+$, and absent US, $\beta^-$ can be entered in PALMS for the RW model to account for certain experimental conditions, such as relative validity \cite{castiello2025symmetrical}.
The RW model's step function algorithm is shown as Algorithm~1 in \hyperref[S1_Appendix]{S1~Appendix}.

\subsubsection{Pearce-Kaye-Hall model}
\label{pearcekayehall}

The Pearce-Kaye-Hall (PKH) model \cite{pearce1982predictive} is a reformulation of the original Pearce and Hall model \cite{pearce1980model}, which provides a formal account for the observed variations in the effectiveness of conditioned stimuli, their associability, in certain Pavlovian conditioning preparations.

The original Pearce-Hall expression yielded undesirable results, such as the associability of non-reinforced stimuli decreasing to zero in a single trial. The version used in this implementation is the revised Pearce-Kaye-Hall rendition \cite{pearce1982predictive}. Another implementation of this model was instantiated in a Java simulator by Grikietis and Mondragón (2016) \cite{PH_simulator}.

The PKH model posits that the effectiveness of a stimulus to enter into association is proportional to the degree of attention it captures. Further, attention is directed towards stimuli whose consequences are unknown. Conversely, attention is reduced when a stimulus consistently predicts a given outcome.

In the PKH model, excitatory and inhibitory learning and predictions are modelled independently, inspired by Konorski's assumption that two different associations between CS and US and a CS and a no-US centre ($\overline{\text{US}}$) are formed in inhibitory preparations \cite{konorski1967integrative}.

The attention to a stimulus $\eks$, $\alpha_\eks$, at each trial $n$ determines the rate of learning and is calculated as follows:

\begin{equation}
\label{alphap}
	\begin{aligned}
		\text{If } \kappa_{\eks}^{n+1}>0, \text{ }\alpha_\eks^{n + 1} &= \gamma \cdot \left|\lambda^n - \VNet{}_i^{n} \right| + (1 - \gamma) \cdot \alpha_\eks^n \\
	\end{aligned}
\end{equation}

Where $\VNet{}_i^{n}$ represents the net outcome prediction at trial $n$, that is, the difference between the cumulative strength of the CS-US associations and the cumulative strength of the CS-$\overline{\text{US}}$ (no-US centre) connections or anti-associations \cite{konorski1967integrative}:
\begin{equation}
\label{vnet}
	\begin{aligned}
		\VNet{}_i^{n} &= \textstyle\sum_i{V_i^{n}}\cdot\kappa_{i}^{n+1} - \textstyle\sum_i\overline{V_i^{n}}\cdot\kappa_{i}^{n+1} \\
	\end{aligned}
\end{equation}

A parameter $\gamma$ controls how much the learning rate change of a stimulus, $\alpha_\eks$, depends on the immediately preceding trial ($\gamma=1$, previous rate), or on early trials ($\gamma \approx 0$).

Following Konorski's assumption \cite{konorski1967integrative}, two different formulations for excitatory and inhibitory learning are posited, an excitatory $\Delta V_\eks$ and inhibitory $\Delta \overline{V_\eks}$ control learning updates.

A variable $\rho$ is used to adjudicate whether the change in associative strength should be computed as excitatory or inhibitory learning. Calculations are defined as follows:

\begin{equation}
	\begin{gathered}
		\begin{aligned}
			\Delta V_\eks^{n + 1} &= S \cdot \beta^+ \cdot \alpha_\eks \cdot \lambda^n & \text{if }  \kappa_{\eks}^{n+1}>0, \text{ } \rho^n \geq 0 \\
			\Delta \overline{V_\eks^{n + 1}} &= S \cdot \beta^- \cdot \alpha_\eks \cdot \left| \rho^n \right| & \text{if } \kappa_{\eks}^{n+1}>0, \text{ }\rho^n < 0 \\
		\end{aligned}\\
        \rho^n = \lambda^n - \left( \textstyle\sum_i V_i^n \cdot \kappa_{i}^{n+1} - \textstyle\sum_i \overline{V_i^n}\cdot \kappa_{i}^{n+1} \right)
	\end{gathered}
\end{equation}

Finally, the net associative strength is estimated according to the equation below:
\begin{equation}
	\begin{aligned}
\VNet{}_\eks^{n + 1} &= V_\eks^{n + 1} - \overline{V_\eks^{n + 1}}
\end{aligned}
	\label{pkh_equation}
\end{equation}

The model's step function algorithm is shown as Algorithm~2 in \hyperref[S1_Appendix]{S1~Appendix}.

\subsubsection{Mackintosh extended model}

\newcommand{\longlambdan}{\rlap{$\;\lambda^n$}\phantom{\left|\rho^n \right|}}
\newcommand{\longI}{\rlap{\textsuperscript{I}}\phantom{\textsuperscript{E}}}

Although the role of attention in associative learning had been previously explored \cite{trabaso,Zeaman}, none had been as influential as Mackintosh's 1975 approach \cite{mackintosh1975theory}. The model proposed a rationale that tied attention to prediction. Thus, in a learning scenario, attention would be focused on the most effective predictors of the outcome.

The Mackintosh Extended (ME) model \cite{lepelley2004} provided a computational rendition of the original formulation.

The ME model posits that attention to stimuli is altered by learning, the speed of which is, in turn, modulated by attentional processes. Accordingly, attention is directed to the best available predictors of a given outcome. Whereas Mackintosh's approach \cite{mackintosh1975theory} merely indicated the direction of change in learning rates, Le Pelley's rendering included a set of equations to quantify it. The new formulation also managed to circumvent some of the obstacles imposed by the original model, which lacked a summation term, making it challenging to account for, e.g., conditioned inhibition \cite{lepelley2004}, which other attempts to modify the original model fail to resolve \cite{moore1985antiassociations,schmajuk1985real}.

Similar to PKH and Pearce and Hall's original proposals, the model hypothesises separate excitatory and inhibitory learning processes. A parameter $\rho$ is calculated to establish which type of association is updated, $\Delta V$ for excitatory CS-US links or $\overline{V}$, for inhibitory CS-$\overline{\text{US}}$ connections, according to the following equations:

\begin{equation}
\label{Eqrho}
\begin{gathered}
	\begin{aligned}
		\Delta V_\eks^{n + 1} &= \alpha_\eks^n \cdot \beta^+ \cdot (1 - V_\eks^n + \overline{V_\eks^n}) \cdot \left| \rho^n \right| & \text{if }\kappa_{\eks}^{n+1}>0, \text{ } \rho^n \geq 0 \\
		 \Delta \overline{V_\eks^{n + 1}} &= \alpha_\eks^n \cdot \beta^- \cdot (1 - \overline{V_\eks^n} + V_\eks^n) \cdot \left| \rho^n \right| & \text{if }\kappa_{\eks}^{n+1}>0, \text{ } \rho^n < 0\\
         \end{aligned}\\
         \rho^n = \lambda^n - \left( \textstyle\sum_i V_i^n \kappa_{i}^{n+1}- \textstyle\sum_i \overline{V_i^n} \kappa_{i}^{n+1}\right)
\end{gathered}	
\end{equation}

This model conceptualises two parameters, $\theta^E$ and $\theta^I$ (also referred to as $\theta^+ and \theta^-$), to modulate the change in attention (or associability) for excitatory and inhibitory updates, which are computed respectively for each present stimulus, that is, if $\kappa_{\eks}^{n+1}>0$, as follows:

\begin{equation}
\label{alpham}
	\alpha_\eks^{n + 1} =
	\begin{cases}
		\alpha^n_\eks -\theta^E \cdot \left( \left| \longlambdan - V_\eks^n + \overline{V_\eks^n} \right|- \left| \longlambdan - \textstyle\sum_i^{i \neq \eks} \left( - V_i^n + \overline{V_i^n} \right) \right| \right) & \text{if } \rho^n > 0 \\[1ex]
		\alpha^n_\eks -\theta\longI{} \cdot \left( \left| \left| \rho^n \right| - \overline{V_\eks^n} + V_\eks^n \right| - \left| \left| \rho^n \right| - \textstyle\sum_i^{i \neq \eks} \left( - \overline{V_i^n} + V_i^n \right) \right| \right) & \text{if }\rho^n < 0
	\end{cases}
\end{equation}

Two further conditions are established: 1) $\theta^E>\theta^I$, to sustain a high associability of the excitor across trials, and 2) at any given trial, $\alpha^n_\eks$ is bounded between 0.05 and 1.

Like in the previous model, the net associative strength is obtained by subtracting the CS inhibitory strength from the excitatory.

\begin{equation}
	\VNet{}_\eks^{n + 1} = V_\eks^{n + 1} - \overline{V_\eks^{n + 1}}
	\label{VNetMack}
\end{equation}

Mackintosh Extended model's step function algorithm is shown as Algorithm~3 in \hyperref[S1_Appendix]{S1~Appendix}.

\subsubsection{Le Pelley with hybrid attention (LPH)}
\label{le_pelley_hybrid}
Le~Pelley \cite{lepelley2004} introduced a hybrid attentional model of learning that aims to accommodate PKH \cite{pearce1980model,pearce1982predictive} and Mackintosh's \cite{mackintosh1975theory,lepelley2004} discordant conceptualisations of attention in learning. The extant conflicting empirical results supporting each standpoint underscore the need to reconcile the two models. Le Pelley's rationale assumes that each approach may be describing different properties. Accordingly, Mackintosh's attention may be characterised as an "attentional associability" ($\alpha^\text{M}$), whose primary role would be to determine which stimulus should be available for learning. On the other hand, Pearce–Hall's mechanism ($\alpha^\text{H}$) would better denote a sort of "salience associability" rather than an attentional one, that would ultimately determine a stimulus learning rate based on its exposure history.

The process of hybridisation is simple enough. Pearce–Hall $\alpha$ is multiplicatively added into the extended Mackintosh model. With $\rho$ calculated as in \cref{Eqrho}, $\alpha^\text{\!M}$ as in \cref{alpham} and $\alpha^\text{\!H}$ as in \cref{alphap}, learning for each present stimulus ($\kappa_{\eks}^{n+1}>0$) in Le~Pelley's hybrid model is defined as follows:

\begin{equation}
	\begin{gathered}
		\rho^n = \lambda^n - \textstyle\sum_i \left( V_i \cdot \kappa^{n+1}_{i}- \overline{V_i}\cdot \kappa^{n + 1}_{i} \right) \\
		\begin{aligned}
			\Delta V_\eks^{n + 1} &= {\alpha^M_\eks}^n \cdot {\alpha^H_\eks}^n \cdot \beta^+ \cdot (1 - V_\eks^n + \overline{V_\eks^n}) \cdot \left| \rho^n \right| & \text{if } \rho^n \geq 0 \\
			 \Delta \overline{V_\eks^{n + 1}} &= {\alpha^M_\eks}^n \cdot {\alpha^H_\eks}^n \cdot \beta^- \cdot (1 - \overline{V_\eks^n} + V_\eks^n) \cdot \left| \rho^n \right| & \text{if } \rho^n < 0
		\end{aligned}
	\end{gathered}
\end{equation}

\begin{equation}
	\begin{gathered}
		{\alpha^M_\eks}^{n + 1} =
			\begin{cases}
				{\alpha^M_\eks}^n - \theta^E \cdot {\alpha^H_\eks}^n \cdot \left( \left| \longlambdan - V_\eks^n + \overline{V}_\eks^n \right| - \left| \longlambdan - \textstyle\sum_i^{i \neq \eks} \left( - V_i^n \cdot \kappa^{n + 1}_{i}+ \overline{V_i^n} \cdot \kappa^{n + 1}_{i}\right) \right| \right) & \text{if } \rho^n > 0 \\[1ex]
				 {\alpha^M_\eks}^n -\theta{\longI} \cdot {\alpha^H_\eks}^n \cdot \left( \left| \left| \rho^n \right| - \overline{V}_\eks^n + V_\eks^n \right| - \left| \left| \rho^n \right| - \textstyle\sum_i^{i \neq \eks} \left( - \overline{V_i^n}\cdot \kappa^{n+1}_{i} + V_i^n \cdot \kappa^{n+1}_{i}\right) \right| \right) & \text{if } \rho^n < 0
			\end{cases} \\[1ex]
				{\alpha^H_\eks}^{n + 1} = \gamma \cdot \left| \rho^n \right| + \left( 1 - \gamma \right) \cdot {\alpha^H_\eks}^{n}
	\end{gathered}
\end{equation}

\begin{equation}
	\VNet{}_\eks^{n + 1} = V_\eks^{n + 1} - \overline{V_\eks^{n + 1}}
\end{equation}

As in the case of $\alpha$ in the Extended Mackintosh model, $\alpha^\text{\!M}$ is bounded between 0.05 and 1. In addition, the values of $\alpha^\text{\!H}$ are confined between 0.5 and 1.

The pseudocode for the step function algorithm of Le Pelley's Hybrid model is shown as Algorithm~4 in \hyperref[S1_Appendix]{S1~Appendix}.

\subsubsection{MLAB model: a RW model extension with a unified attention rate}

Proposals exist that have explored variable attentional parameters within the RW model to characterise selective attention within its framework, while preserving the fundamental learning algorithm (e.g., \cite{frey1978model,nishimura2020rescorla}) or in more complex real-time implementations (e.g., \cite{luzardo2017rescorla,schmajuk1996latent}). We are adding to this effort by implementing an extension to the original Rescorla-Wagner, the MLAB model, that defines a unified variable attentional learning rate that replaces the constant $\alpha$ of the original formulation. This conceptualisation of $\alpha$ incorporates both the ideas suggested by Mackintosh \cite{mackintosh1975theory} and Pearce and Hall \cite{pearce1980model}. Accordingly, attention to a stimulus decreases with exposure in a negatively accelerated fashion, controlled by a decay constant $d$. In US trials, when $\lambda$ is positive, this decay is counteracted by a value which is a function of the stimulus' predictive value and the prediction error, modulated by the initial $\alpha$ value. That is, both the cue-predictive information and the overall US expectancy, defined as the discrepancy between the current US value and the total global prediction, oppose the decrease in the learning rate. On the contrary, in non-US trials, attentional decay is boosted by the same proportion. The concrete formulation to calculate $\alpha$ is shown in \cref{alphaMLAB} below:

\begin{equation}
\label{alphaMLAB}
	\text{If } \kappa_{\eks}^{n+1}>0,\text{ }\alpha_\eks^{n + 1} =
	\begin{cases}
		\alpha^n_\eks \cdot \left(1-d\right)+ \alpha^0_\eks \cdot V_\eks^n \cdot \left(\lambda^n- \textstyle\sum_i^{}V_i^n \cdot \kappa_{i}^{n+1}\right) & \text{if } \lambda^n > 0 \\[1ex]
		\alpha^n_\eks \cdot \left(1-d\right)- \alpha^0_\eks \cdot V_\eks^n \cdot \left(\lambda^n- \textstyle\sum_i^{}V_i^n \cdot \kappa_{i}^{n+1}\right) & \text{otherwise }
	\end{cases}
\end{equation}

A formal description of the  MLAB model is provided below. A thorough evaluation of this model is beyond the scope of this paper. It is included here as a showcase scenario to illustrate how PALMS can be used to assess model extensions and new developments. In particular, the unified attention formulation proposed here is capable of simulating traditional Pearce and Hall stronghold phenomena, such as latent inhibition \cite{hallpearce1979,Escobar_2003} and Hall and Pearce negative transfer \cite{negative_transfer}, and partial reinforcement effect \cite{swan1988orienting,two_kinds_of_attention}, as well as some of the original \cite{mackintosh1975theory} or extended Mackintosh \cite{lepelley2004} models predictions, such a learned irrelevance \cite{lepelley2004}, or other derived effects, for example Haselgrove and collaborator's \cite{two_kinds_of_attention} Experiment 3.

The extension enables the RW model to account for phenomena that the original rendition cannot, and that cannot be jointly predicted by Mackintosh-like \cite{mackintosh1975theory,lepelley2004} or Pearce and Hall's models \cite{pearce1980model,pearce1982predictive,E_PH_elaboration}.

The MLAB model's step function algorithm is shown as Algorithm~5 in \hyperref[S1_Appendix]{S1~Appendix}.
 
\subsection{Computational implementation and design features}
\label{experimental_methodology}

This section summarises the implementation of PALMS, its characteristic elements, and the design and functionalities of its software instantiation. 

The implemented environment includes a visual interface that resembles abstract standard experimental settings in Pavlovian conditioning, thereby allowing direct input of an entire design comprising one or multiple phases, organised in successive columns, with each phase supporting a specific number of sequential or random trials. Different experimental groups can be entered in parallel rows, enabling independent execution of each group while simultaneously presenting results within the same plot.

\subsubsection{Stimulus representation and trial configuration}
\label{extra_names}\label{trial_computation}

A trial consists of one or more conditioned stimuli (CS) that can be paired with an unconditioned stimulus (US).

A conditioned stimulus is defined by a single letter \textsf{A--Z}, optionally followed by an arbitrary number of prime (\textsf{\textquotesingle}) characters, such as \textsf{A\textquotesingle{}\textquotesingle{}}, or a caret symbol (\tcaret{}) followed by a number, such as \textsf{A\tcaret{}6}.
Both can be combined; for example, \textsf{A\textquotesingle{}\textquotesingle{}\textquotesingle{}\tcaret{}12} is a valid CS that's unrelated to the CS \textsf{A}, or any CS mentioned in this paragraph.

The presence of a US is denoted by the symbol \textsf{+}, and its absence by the symbol \textsf{-}.
A double-strength US can be applied with a double stimulus symbol \textsf{++}.

Following the original theoretical approaches, learning is conceptualised as changes in the associative strength ($\Delta V$) connecting two events, a predictor (CS) and an outcome (US).
At any given learning trial, a CS holds a predictive value over the outcome, defined by the associative strength of the connection, $V$, which is updated with new experience. Prior to experience, the initial prediction in all models is zero.

Models in PALMS are trial-based elemental learning models. Hence, the simulator's learning and prediction algorithms are correspondingly updated for each stimulus and trial.

Compound stimuli (e.g., \textsf{XYZ}) do not undergo learning. Their predictive value is defined as the sum of the associative strengths of their constituent stimuli at a given trial, as shown in \cref{multiple_cs}.

\begin{equation}
	V_{XYZ} = V_X + V_Y + V_Z
	\label{multiple_cs}
\end{equation}
An exception to this summation rule arises when configural cues \cite {r_w_inhibition_pavlovian} are considered to operate in an experiment. These are explained in \cref{configural_cues}.

The following symbols are universal for all models in PALMS.
\begin{enumsymbol}
	\item[$V_\eks$]
		The associative strength (value) between CS \eks{} and the US link.
	\item[$\alpha_\eks$]
		Learning rate of CS \eks{}.
		A constant value representing the stimulus intensity in the RW model, and a variable that represents the changes in the associability or effective stimulus salience with learning in the remaining models.
	\item[$\beta^+$] A parameter representing the effective salience of a present US.
	\item[$\beta^-$] A parameter representing the effective salience of an absent but expected US.
	\item[$\lambda$] The asymptote of learning, which represents the upper limit of prediction, associative strength, that a US can support. This is set to zero when the US is not present.
\end{enumsymbol}

Constants and variables specific to a given theoretical approach will be introduced when the corresponding model is selected.

\subsubsection{Configural cues}
\label{configural_cues}

Configural Cues can be defined for all models. These represent Wagner and Rescorla's intuition that whenever two or more stimuli are presented in compound, some \textit{configural} elements that uniquely represent the conjunction of stimuli are formed. These distinctive cues, which are emergents of the exact combination of stimuli, compete with the compound cues for associative strength. The simulator, if so chosen, computes the associative strength of configural cues; these are included when estimating the global error and added to the calculation of the total compound predictive value \cite {r_w_inhibition_pavlovian}.

The total compound associative value will thus be estimated as 
\cref{multiple_cs_cc}.
Configural cues are represented as \textsf{q(XYZ)}, where the parentheses identify the individual stimuli that form the compound and originate the configuration.

\begin{equation}
	V_{XYZ} = V_X + V_Y + V_Z + V_{ q\left ( XYZ \right) }
	\label{multiple_cs_cc}
\end{equation}

\label{program_design}

\newcommand{\titlebox}[3]{
	\draw[fill = black, fill opacity = 0.6, draw = none] (#1) rectangle ++(#2, -10pt);
	\node[
		anchor = north west, align = center, minimum width = #2,
		font = \tiny\itshape, text = red] at (#1) {\scalebox{0.75}{#3}};
}

\newcommand{\infobox}[4]{
	\draw[fill = black, fill opacity = 0.6, draw = none] (#1) rectangle ++(#3, -10pt);
	\node[
		anchor = north west, align = center, minimum width = #3,
		font = \tiny\itshape, text = red
	] at (#1) {\scalebox{0.75}{#4}};
	\draw[thick, red] (#1) rectangle (#2);
}

\newcommand{\partfont}[1]{\textsf{#1}}
\subsubsection{Layout design}
In this section, we present the layout of PALMS' visual interface. We first introduce the layout sections and portray the design input and optional functionalities. We later describe the command-line interface.

\begin{figure}[ht!]
\centering
\includegraphics[width=\linewidth]{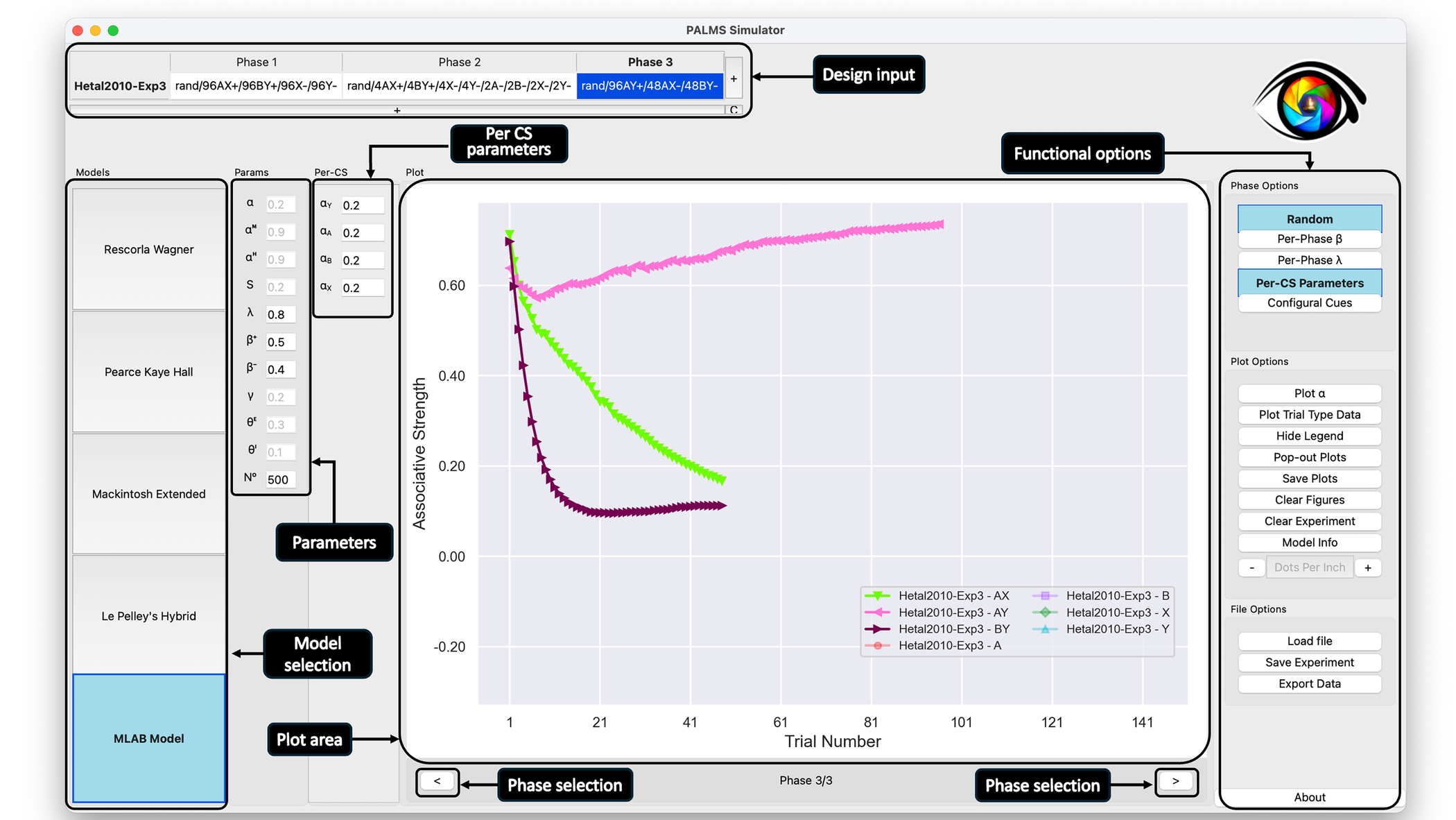}

	\caption{PALMS layout displaying a MLAB model simulation of Haselgrove et al., Experiment 3 \cite{two_kinds_of_attention}. The design includes three Phases and one Group. The different sections of the layout are identified and framed by a black rounded rectangle, namely, the Design input (top of the visual interface), and from left to right, the Model selection buttons, model Parameters area, Per CS Parameters area, the Plot area, the simulation Functional options section, and at the bottom, the Phase selection arrows.}

	\label{big_example}
\end{figure}

The implementation's graphical interface is displayed in \Cref{big_example}. The figure shows a simulation of the MLAB model, explained in \cref{le_pelley_hybrid}.
The experimental design follows Haselgrove et al. (Experiment 3, \cite{two_kinds_of_attention}) and comprises three phases. Unlike in the original experiment, the second phase of the simulation corresponds to the test trials included at the end of Phase 1 in the empirical version, which involved specific trial block sequences. In the simulation, these were presented randomly, in a separate phase. Thus, Phase 3 in the simulation matches the empirical final test. To expand the implementation's functionality, the interface includes a section that allows the input of individual $\alpha$ values per CS.

PALMS supports the simulation of different models described in \cref{adaptive_types}. The interface allows quick model selection by clicking the corresponding buttons in the \partfont{Models} selection area. This action will be followed by the immediate computation of the model's predictions.  

Experimental designs are entered in the Design input table at the top of the interface.
Each row of the table represents one group, which runs independently.
Each column represents one phase, which runs sequentially.
Groups can be renamed by double-clicking on their names.
This way, meaningful names can be entered for each group that will appear in the legend of the plot.

The simulation for the selected model starts automatically as soon as a phase is entered, using the default model parameters displayed in the \partfont{Params} section. These values can be directly modified. Changing the experiment or any parameter automatically initiates a new simulation. Some parameters are model-specific and will be disabled when not used.

A figure depicting associative values per trial is displayed in the \partfont{Plot} area. Data will be updated automatically as the design simulation is completed. A stimulus or compound values can be removed from view by clicking the figure legend or a line. Switching between phases can be accomplished by pressing the right or left arrow keys below the plots.

Initial parameter values can be defined separately for each CS in the \partfont{Per-CS} panel, which is opened by clicking the corresponding button in the functional options layout. \partfont{Phase Options} can be altered by clicking one of the buttons in the right section. For instance, random trial presentations can be selected, highlighting one or multiple phases and groups, and clicking the \textsf{Random} button or simulating different $\beta$ and $\lambda$ values for each phase by pressing the corresponding buttons. When using compound stimuli, \textsf{Configural Cues} can be enabled for all models. The use and calculation of these will be explained later.

 A number of \textsf{Plot Options} are also provided. For instance, the associative strength, or $\alpha$ values, can be displayed alternatively.  Selecting \textsf{Plot Trial Type Data} allows the independent visualisation of, for instance, the associative values of a stimulus \textsf{A} during \textsf{A+} trials and \textsf{AB-}, or \textsf{A-} trials. These options aim to improve direct comparison between simulated results and experimental observations. The \textsf{Pop-out Plots} button opens figures in an editing window, allowing the data to be saved. Alternatively, plots can be saved directly by choosing the \textsf{Save Plots} option, which opens a menu that allows modifying the plot size and saving legends in a separate file, useful for visualising large designs. Figures can be cleared. When this option is selected, the legend remains to allow one-by-one selection of cues. In addition, a design can be fully reset by pressing \textsf{Clear Experiment}.

 The mathematical formulation for each model is rendered in \LaTeX{} and appears when pressing the \textsf{Model Info} button.

 Experiments can be saved to and loaded from \texttt{.rw} files, which follow the specification provided in \cref{rw_file} using the \textsf{Save File} and \textsf{Load Experiment} buttons.

\subsubsection{Experimental design description}
\label{experiment_table}

By default, table cells for two phases and two groups are displayed when PALMS is opened. Multiple phases and groups can be added by pressing the vertical and horizontal \framebox{\small +} buttons in the Design input table; empty groups and phases can be deleted with the \fbox{\small C} button.

Simulations are independent across groups; that is, the results in one group do not affect those in other groups. In contrast, simulations are run sequentially across phases, so that results from one phase affect the subsequent phase. Results and properties for each conditioned stimulus on a given phase, such as associative strength $V$ and learning rate $\alpha$, are stored and carried over to the next phase.

On any given \textsf{Phase}, trials are defined according to the structure below:
\graybox{
    \ttfamily
	\small
	\begin{tabular}{r @{\hspace{1ex}} c @{\hspace{1ex}} l}
		Phase & := & [rand/] [beta=$\beta$/] [lambda=$\lambda$/] T\textsubscript{0}Trial\textsubscript{0} [/ T\textsubscript{1}Trial\textsubscript{1}[/\ldots{}]] \\
		Trial & := & CS\textsubscript{0}[CS\textsubscript{1}[\ldots{}]] US \\
		\phantom{C}T & := & \{0-9\}* \\
		CS & := & \{A-Z\} | \{A-Z\}\textquotesingle{}+ | \{A-Z\}\textquotesingle{}*\tcaret{}\{0-9\}+ \\
		US & := & \{++,+,-\}
	\end{tabular}
    \label{trial_description}
}

A \textsf{Trial} is defined as a set of conditioned stimuli followed by the presence (\textsf +) or absence (\textsf -) of an unconditioned stimulus.
Using a double strength US \textsf{++} is equivalent to doubling the $\beta^+$ parameter in this particular phase.

Conditioned stimuli are represented as a single letter from \textsf A to \textsf Z.
Complex experiments that involve a large number of CSs can also be entered by appending an arbitrary amount of \textsf{\textquotesingle{}} characters (such as \textsf{A\textquotesingle{}\textquotesingle{}}) or a \textsf{\tcaret{}} character followed by a number (such as \textsf{A\tcaret{}14}) to each letter of the alphabet.
CSs, as defined above, are treated as independent of one another.

Compound stimuli are computed by summing the associative strengths of multiple CS within a trial.

Trials within a phase are separated by a slash \textsf{/}. A number of identical trials, e.g., \textsf{A+/A+/A+}, can be input in any given sequence.
Identical trials are encoded as unique trial types.
Thus, \textsf{A+/A+/A+} is equivalent to \textsf{3A+}.
The simulator computes the values in the given order, e.g., \textsf{3A+/2B+/2A+}, but values are reported per trial type.

The process by which sequential trials are computed in PALMS is shown as Algorithm~6 in \hyperref[S1_Appendix]{S1~Appendix}.

\subsubsection{Experiment options}

PALMS provides several options to simulate a wide range of associative learning experiments.

\textbf{Randomised phases.} By default, PALMS runs the trials in each phase sequentially in the exact order they appear in the design table.

The \textsf{Random} button adds a \texttt{rand/} prefix to the currently selected phase, which imposes a randomisation of trials within the phase with a certain number of iterations, specified by the \textnumero{} parameter, at the bottom of the \partfont{Params} section.
The resulting associative value of a CS at each trial is the average of the sequence's value for that stimulus and trial order across all runs, disregarding its specific location within each sequence. For example, random presentations of \textsf{10A+/10AX-}, outputs 20 $V_A$ values, each corresponding to the average of each consecutive \textsf{A}, regardless of whether it appeared on a \textsf{A+} or \textsf{AX-} trial.

Pseudocode for the randomisation and averaging is shown as Algorithm~7 in \hyperref[S1_Appendix]{S1~Appendix}.

\textbf{Per-phase $\beta$ and $\lambda$ parameters.} The \textsf{Per-phase $\beta$} and \textsf{Per-phase $\lambda$} buttons add the \texttt{beta=/} and \texttt{lambda=/} prefixes to the currently selected phase, respectively.
This overrides the default $\beta$ and $\lambda$ parameters for the trials of a single phase. This is a useful way to simulate the effects of a stronger or weaker US.

Similarly, using \texttt{++} as a US doubles the $\beta$ parameter.

\textbf{Compound stimuli and configural cues computation.} In \cref{trial_computation} we state that, according to standard elemental theories of associative learning, the strength of a compound stimulus is calculated as the sum of the values of the individual stimuli that compose it.
However, if the \textsf{Configural Cues} button is selected, a configural cue for each compound stimulus in the experiment is calculated as defined by Wagner and Rescorla \cite{r_w_inhibition_pavlovian}.

Configural cues are treated as additional stimuli, emergent properties of specific stimulus combinations.
Thus, the configural cue of a compound stimulus \textsf{AB} is distinct from the configural cue of \textsf{ABC}.
Unlike compound stimuli, configural cues are learned about and compete to gain associative strength with all other present stimuli.

We enable this capability with the \textsf{Configural Cues} button, which selects a configural cue for each compound stimulus in the experiment.

Configural cues are identified by the letter \textsf{q} followed by the compound stimuli in brackets; for example, the configural cue of \textsf{AB} is \textsf{q(AB)}.

Configural cues have the same properties as every other CS and act as a CS in every way.
By default, the configural cue assigned $\alpha$ equals the product of the constituent CS values. The default value can, however, be modified by selecting the \textsf{Per-CS Parameters} button.

The configural compounds' associative strength is then calculated as the sum of the component stimuli's value and the configural cue associative strength, as described in \cref{configural_cues}.

\subsubsection{Plotting options}

\textbf{Plotting the learning rate $\alpha$.} As an alternative to plotting the associative strength of different stimuli, PALMS can plot the learning rate $\alpha$ for each trial.
This capability also applies to the options \textsf{Pop-out Plots} and \textsf{Save Plots}, which save the stimuli's learning rate rather than their associative strength.

\textbf{Trial type data.} Separate trial-type associative strength values can be displayed alongside per-CS and compound values. These are extracted from the main computation by selecting \textsf{Plot Trial Type Data} and allow for isolating stimulus, or compound-level predictions, from the individual process of cue learning, facilitating comparisons with real behavioural data.

\textbf{Legend options.} The plot legend in PALMS supports a high degree of interactivity, which is essential for working effectively with experiments involving a large number of stimuli.

The legend can be hidden and displayed with the \textsf{Hide Legend} toggle.
When the legend is shown, users can also drag it to change its location.

More significantly, if there is a large amount of stimuli present in the experiment, the legend uses pagination to facilitate visualisation within the screen.
When saving plots, the legend can be saved separately with the ``Separate Legend'' option.

\textbf{Pop-out plots.} After an experiment is plotted into the main PALMS window, these same plots can be popped out into the default Matplotlib figure viewer, which allows basic editing and saving.
These plots are essentially the same as those in the main interface.

\textbf{Save plots.} The plot images can be saved with the \textsf{Save Plots} button, which saves one plot per phase with names \texttt{file\_1.png}, \ldots, \texttt{file\_n.png}.

The dimensions of the figures can be specified separately.
Additionally, the legend can be added in a separate file \texttt{file\_legend.png}; this is useful when a large number of stimuli will not fit into the main plots.

\textbf{Hiding conditioned stimuli.} Lines corresponding to CS in certain experiments can be hidden and restored from the main plot by clicking the lines from either the legend or the plot itself.

\Cref{big_example} shows an example of this: the stimuli representing the CS \textsf{B}, \textsf{X}, and \textsf{Y} are hidden from view and only visible from the legend.

\textbf{Saving and loading experiments.}
\label{rw_file}

Experiments can be saved to and loaded from a \texttt{.rw} file.

Data can be saved by pressing the \textsf{Save Experiment} button on the graphical interface, which produces a CSV-style table file with the pipe character \texttt{|} as the separator. This file is separated into a parameters section and an experimental section.
Lines starting with \texttt{@} contain the parameters that are used for the experiments; parameters that do not have a value here will use a default value. The table contains one column per phase and one row per group.

\begin{plisting}[caption=Example of saved \texttt{.rw} file]
@model=Le Pelley's Hybrid
@lambda=0.7;beta=0.6;betan=0.5;gamma=0.30;thetaE=0.4;thetaI=0.2
@alpha_D=0.1;alpha_mack_D=0.3;alpha_hall_D=0.7
Novel|5B+/5C-/5D-||rand/beta=4/5A+/5C-/5D-
NegTransfer|5A+/5C-/5D-||rand/beta=4/5A+/5C-/5D-
Change|5A+/5C-/5D-|rand/2A-/2C-/2D-|rand/beta=4/5A+/5C-/5D-
\end{plisting}

\textbf{Exporting data.}
\label{export_data}

The \textsf{Export Data} button exports the result of the current experiment into a CSV file with data equivalent to the plot being shown, while also including extra data such as the inhibitory and excitatory strengths of each stimulus.

This feature is useful for comparing, contrasting, and processing the results of various experiments.
The data can also be opened in a spreadsheet such as Microsoft Excel.

\subsubsection{Command-line interface}
\label{cli}

In addition to the graphical user interface (GUI), PALMS provides a command-line interface that can be run from the command line.
This functionality can be useful for automating multiple tests, running tests with a large number of stimuli, and for quick testing without requiring the full GUI.

The interface can be accessed by running \texttt{python PALMS.py cli [commands]} on the terminal.
This command takes a set of optional arguments and one single optional positional argument, \texttt{experiment\_file}, which contains an \texttt{.rw} file with an experiment.

By default, the CLI pops up one plot per phase of the data in the \texttt{.rw} file.
The \texttt{-{}-print-results} option can be used to print the results of the experiment to stdout. Results can be saved to a file, instead, using the \texttt{-{}-save-results} option.
This last command is equivalent to the \textsf{Export Data} button in the GUI, presented in \cref{export_data}.

The output plots can be saved with \texttt{-{}-savefig [filename.png]}, which saves the figure to \texttt{\allowbreak filename\_1.png}, \ldots{}, \texttt{\allowbreak filename\_n.png}.
Additionally, \texttt{-{}-singular-legend} removes the legend from the plots and instead saves it to a separate \texttt{\allowbreak filename\_legend.png} file.
This is useful when plotting large experiments.

An overview of every command, displayed when run with the \texttt{-{}-help} option, can be found in \hyperref[S2_Appendix]{S2~Appendix}.

\subsubsection{Releases and source code}

\textbf{Main release.} We provide up-to-date compiled releases for Windows, MacOS, and Linux under \url{https://github.com/cal-r/PALMS-Simulator/releases/tag/latest}. The compiled release file corresponding to the desired operating system can be downloaded and run \texttt{PALMS.exe} on Windows systems, \texttt{PALMS.app} on macOS systems, or the executable \texttt{PALMS} on Linux directly.

An additional \texttt{Experiments/} directory is also present in all releases, which contains the design files of the experiments used in this paper.

Due to the size of the libraries bundled in this executable, it might take some time to run the first time it is opened.
The MacOS executable is notarised and certified to run without requiring authorisation, but Windows users may need to manually approve execution.

\textbf{Source code.} The source code of the implementation is released on the following public GitHub repository: \url{https://github.com/cal-r/PALMS-Simulator}.

The program can be run using Python $\geq$ 3.10 along with some dependencies listed under \texttt{\allowbreak requirements.txt} and available under \texttt{pip} virtual environments.

\begin{plisting}[caption=Running instructions with custom virtual environment]
$ source ~/venv/bin/activate
$ pip install -r requirements.txt
$ python PALMS.py#$
\end{plisting}

The code is released under the GNU Lesser General Public Licence\cite{lgpl}, which allows researchers to use, study, and modify the software for any purpose. 

Users may also distribute the original and modified versions, provided that the LGPL-covered versions remain under this licence and that the authors are credited.

\textbf{Adding new models.} The code in PALMS is laid out in an intuitive way that makes it simple for a researcher to add custom models to the current implementation.

The file \texttt{Models.py} contains a \texttt{Model} abstract base class that can be sub-classed into different models.
The class also contains a \texttt{types} class method that can be modified to return any new model created by the researcher.

New models are defined as new classes which subclass the \texttt{Model} class and can contain the following methods.

\Cref{model_interface} shows the abstract methods that can be overloaded by new models.

\begin{plisting}[style=py, label=model_interface, caption=Interface for new model classes added to \texttt{Models.py}]
# Run a step of a certain adaptive type. This is the only function that
# requires being overloaded by subclasses of Model.
#   s: Stimulus, definition of the stimulus at a certain point (see Environment.py).
#   rp: RunParameters, parameters passed to the model.
def step(self, s: Stimulus, rp: RunParameters)

# List of parameters enabled by this model. Parameters not enabled will
# be marked as gray on the GUI. By default, enable all parameters.
def parameters(cls) -> list[str]:

# Dictionary of default values for certain parameters.
# If these parameters are not changed manually, then when changing
# to this model the parameter in each key will take the form of the value.
def defaults(cls) -> dict[str, float]:

# Dictionary of bounds for certain parameters.
# If a parameter is in the dictionary, the GUI will warn when its value is outside
# the [min, max] bounds returned by this function.
def bounds(cls) -> dict[str, tuple[float, float]]:
\end{plisting}
 
\section{Results}
\label{results}

In this section, we present simulated results from several experiments and models to illustrate the models' computational implementation and operation capabilities.

\subsection{Blocking and stimulus salience}

The phenomenon of blocking refers to a failure to learn (or express \cite{stout2007sometimes}) a stimulus-outcome relationship when the target predictor is accompanied by another stimulus that has already become a reliable signal of the outcome.

Although perhaps the most archetypal phenomenon in associative learning, blocking has not been devoid of diverse interpretations. The most common of them is the one derived from RW's global error-correction mechanism \cite{RescorlaWagner}. From this perspective, learning about the new cue and the outcome is impaired (blocked) when the latter is already predicted by other present stimuli. Because the outcome is expected, there is very little or no prediction error to drive further learning. In other words, for the RW model and others that include a global error term \cite{lepelley2004,pearce1980model,brandon2003stimulus,Wagner1981,kokkola2019double}, blocking results from cues competing to gain associative strength.

A different approach was proposed by Mackintosh \cite{mackintosh1975theory}, who posited that blocking stems from attentional biases; learners would adaptively shift their attention away from cues that are redundant or poor predictors of an outcome.

We simulated a within-subjects blocking design \cite{blaser2008within,prados2013blocking,rescorla1981within}, using two models: Rescorla-Wagner and Mackintosh Extended. Unlike Mackintosh's postulates \cite{mackintosh1975theory}, the ME model does include a global error term. However, the model's attentional variability follows the original rules proposed by Mackintosh. Comparing the results from both the Rescorla-Wagner and Mackintosh Extended models may provide researchers with useful insights into the contribution of each type of explanation to the phenomenon.

More specifically, we employed the procedure described in McNally and Cole \cite{mcnally2006opioid}. In their Experiment 1, during Phase 1, rats received 3 days of 4 conditioning trials each in which a stimulus \textsf{A} was followed by a mild electric shock. During the next two days of Phase 2, they received two presentations of \textsf{AB} and two of \textsf{CD}, each followed by the shock. In Phase 3, 4 presentations of \textsf{B} and \textsf{D} were given. Freezing behaviour was measured. We simulated this experiment in Group \textsf{Blk Exp1 McN}, with random presentations of the stimuli in the last two phases.

Empirical research suggests that salient cues are less likely to be blocked \cite{denton2006attention,hall1977loss}. Thus, to better assess PALMS's capabilities and to explore the models' potential to account for such salience effects, Group \textsf{Blk HS Target} was added to the design. Simulation conditions in the latter were identical to the previous, except that the salience of the stimulus to be blocked (target) \textsf{B} was increased, and, to match the overshadowing effect in both experimental conditions, the salience of the control cue \textsf{D} was matched to \textsf{B}'s. To keep a consistent stimulus labelling, the same letters were used, marked with a prime symbol to distinguish them from those in Group \textsf{Blk Exp1 McN}. The design and parameters for these simulations can be seen in \cref{tab:Exp1-within-design,tab:Exp1-within-params}.

\begin{table}[ht!]
    \centering
    \caption{Within-subjects blocking design for Simulation Set 1}
    \label{tab:Exp1-within-design}
    \sffamily
    \footnotesize
    \begin{tabular}{l l l l}
        \toprule
        & Phase 1 & Phase 2 & Phase 3 \\
        \midrule
        Blk Exp1 McN & 12A+ & 4AB+/4CD+ & 4B\tminus{}/4D\tminus{} \\
        Blk HS Target & 12A'+ & 4A'B'+/4C'D'+ & 4B'\tminus{}/4D'\tminus{} \\
        \bottomrule
    \end{tabular}
\end{table}

\begin{table}[ht!]
    \centering
    \caption{Simulation parameters for Simulation Set 1}
    \label{tab:Exp1-within-params}
 
    \smallgraybox{\hspace{-1.5ex}
        \begin{tabular}{>{\raggedright\arraybackslash}m{.08\columnwidth} >{\raggedright\arraybackslash}m{.88\columnwidth}}
            \textbf{RW} & $\alpha_{A,B,C,D} = 0.15$ \; $\alpha_{A',C'} = 0.15$ \; $\alpha_{B',D'} = 0.30$ \; $\lambda = 0.8$ \; $\beta^+ = 0.5$ \; $\beta^- = 0.3$ \\
            \textbf{ME} & $\alpha_{A,B,C,D} = 0.15$ \; $\alpha_{A',C'} = 0.15$ \; $\alpha_{B',D'} = 0.30$ \; $\lambda = 0.8$ \; $\beta^+ = 0.5$ \; $\beta^- = 0.3$ \; $\gamma = 0.1$ \; $\theta^E = 0.25$ \; $\theta^I = 0.2$
        \end{tabular}
    }
\end{table}

\begin{figure}[ht!]
    \centering
\includegraphics[width=\linewidth]{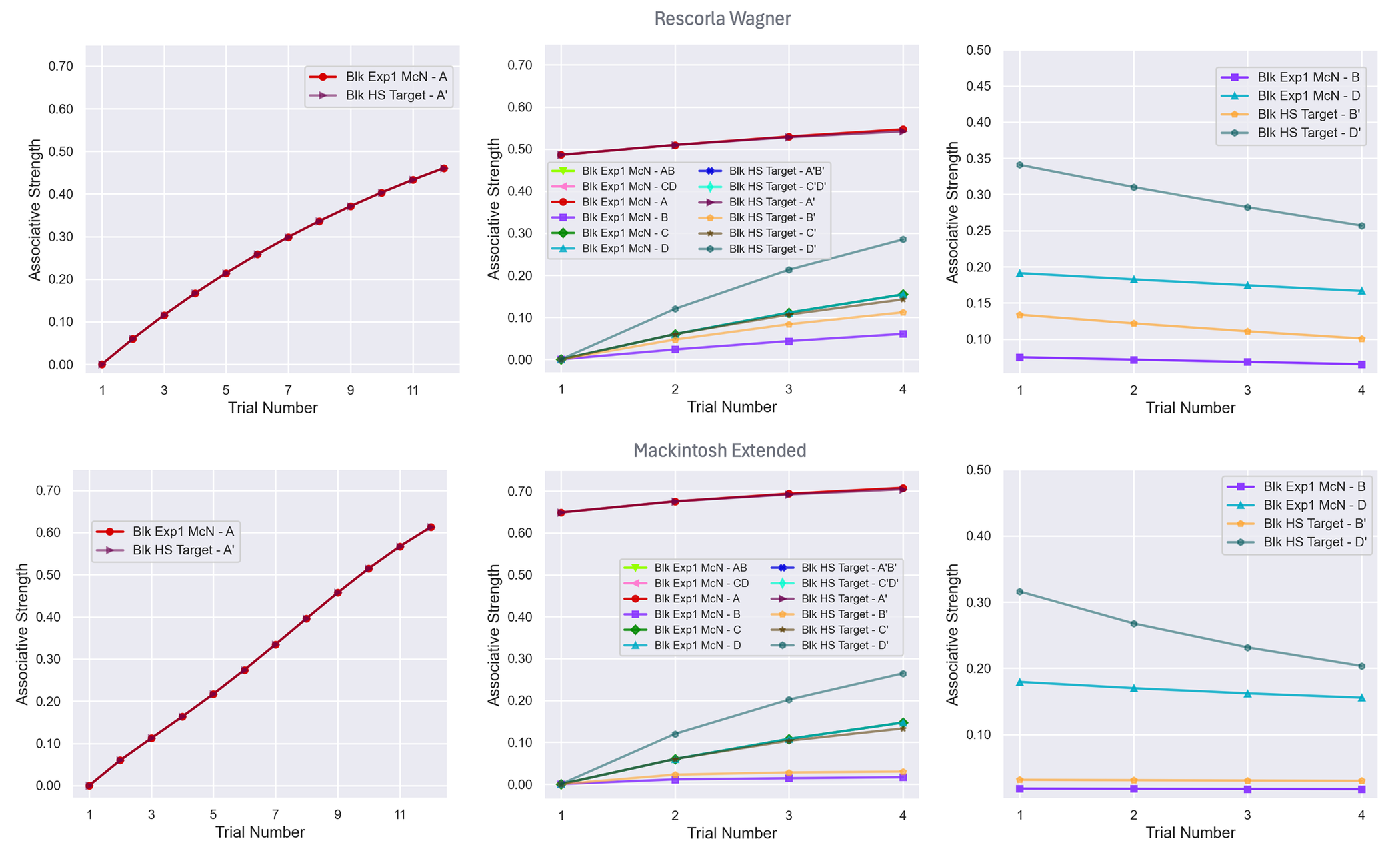}
\caption{Simulation results for a within-subjects blocking design for Groups Blk Exp1 McN and Group Blk HS Target across phases. From left to right, Phase 1, Phase 2, and Phase 3. Individual stimuli are shown with different coloured lines and marker shapes. The top panel display RW simulations, whereas the bottom panel presents Mackintosh Extended simulations of the same hypothetical experiment. Random trials included 500 sequences.}
\label{fig:Fig2-Blocking}
\end{figure}

RW predicted associative strength values per phase are shown in \cref{fig:Fig2-Blocking}. From left to right, each panel displays the associative strength of the individual stimuli. In Phase 1, acquisition develops identically for \textsf{A} and \textsf{A'}, with lines overlapping. During Phase 2, compound conditioning, cues \textsf{A} and \textsf{A'} continue to acquire associative strength, whereas conditioning to \textsf{B} and \textsf{B'} remained low in comparison. Conditioning to \textsf{C}, \textsf{C'}, and \textsf{D} began to develop at a similar rate. Finally, conditioning to \textsf{D'} developed faster than to \textsf{C'}. The simulated pattern of results during Phase 3 in Group \textsf{Blk Exp1 McN} replicated McNally and Cole's empirical data \cite{mcnally2006opioid}. That is, like in their experiment, the strength of the association between the stimulus \textsf{B} and the outcome was weaker than between \textsf{D} and the outcome. This difference is considered indicative of a blocking effect. According to the RW model, stimulus \textsf{A} would have blocked the acquisition of \textsf{B} in Phase 2, in comparison to that acquired by \textsf{D}, whose compound partner stimulus had not received previous training.

Results in Group \textsf{Blk HS Target} are more interesting for our purposes. The increase in salience to the target stimulus \textsf{B'} and to the corresponding control \textsf{D'} resulted in an apparent reduction of the blocking effect when the two target stimuli \textsf{B} and \textsf{B'} were compared across groups. The associative strength of stimulus \textsf{B'} was higher in comparison to that of \textsf{B} in Group \textsf{Blk Exp1 McN}. However, the associative strength of stimulus \textsf{D'} also showed a considerable boost in comparison to the strength of \textsf{D} in Group \textsf{Blk Exp1 McN}. Thus, comparatively, the difference in associative strength between \textsf{B} and \textsf{D} and between \textsf{B'} and \textsf{D'} remains high, indicating that the relative magnitude of blocking in \textsf{Blk HS Target} remains high.

Mackintosh Extended simulations produce equivalent results. However, the attenuation of the blocking effect, shown by the target stimulus' associative strength in Group \textsf{Blk HS Target}, was considerably less marked, almost non-existent, than that predicted by RW. Relative to the control stimulus \textsf{D'}, the predictive value of which significantly increased, the ME simulation points to an enhanced blocking effect.

This result suggests that first, different salience overshadowing between stimuli in the control condition plays a determinative role, which should be considered when assessing the reliability of blocking in within-subjects designs. More critically, the difference in magnitude of the attenuation of the blocking effect to the target cue B' in Group \textsf{Blk HS Target} predicted by the ME model relative to that predicted by RW suggests that additional attentional factors may play a determinative role in the observation of the blocking effect. According to the ME model, the blocking stimulus \textsf{A'} will undergo an increase in associability in Phase 1, which may counteract the effect of the increased salience of the blocked cue. Please note that alpha values have been omitted to save space; readers can run the experiment included in the software release to verify the accuracy of the statements. In addition, the increased salience of its control \textsf{D'} will boost acquisition, resulting in a net enhancement of the blocking effect in within-subjects preparations.

In conclusion, this implementation enables researchers to conduct a meticulous analysis of the distinct predictions made by these two models. It facilitates a thorough evaluation of the influence of individual parameters and the observation of contrasting predictions, which may help them refine their experiments and theories.

\subsection{Learned irrelevance}
The phenomenon of learned irrelevance has generated abundant empirical research \cite{aberg2020neurocomputational,myles2023latent} and theoretical debate. First reported by Mackintosh \cite{mackintosh1973stimulus}, the nature of the phenomenon has been the subject of ongoing discussion, with researchers arguing that learned irrelevance is a phenomenon in its own right \cite{baker1979preexposure,matzel1988learned} and others deeming it as a confounding resulting from the combined effects of latent inhibition and the so-called US preexposure effect \cite{bonardi1996learned},
\cite{bonardi2003learned}.

Contrary to the RW prediction, which posits that uncorrelated exposure between a CS and US leads to a net zero associative strength, empirical results suggest that something else is learnt, capable of delaying both excitatory and inhibitory subsequent learning \cite{bennett2000context}. A process of outcome-specific associability loss has been suggested as responsible for the effect, which would nevertheless be subject to generalisation to similar outcomes \cite{le2003learned}.

To test some of these ideas and the capabilities of his model, Le Pelley \cite{lepelley2004} simulated a simplified learned irrelevance design. We are using the same design here. The exact simulation input is shown in \cref{tab:Exp2-LIrrev}, the parameters are listed in \cref{tab:Exp2-LIrrev-params}, and the trial order in Phase 1 was randomised. We compared the predictions from LPH model with those from the Pearce-Kaye-Hall, Mackintosh Extended, and MLAB models.

\begin{table}[ht!]
    \centering
    \caption{Le Pelley, 2004 learned irrelevance design for Simulation Set 2}
    \label{tab:Exp2-LIrrev}
    \sffamily
    \footnotesize
    \begin{tabular}{l l l}
        \toprule
        & Phase 1 & Phase 2 \\
        \midrule
        Learned Irrelevance & 20X+/20X\tminus{}/20AX+/20AX\tminus{} & 8A+ \\
        CS-preexposure & 20X\tminus{}/20X\tminus{}/20AX\tminus{}/20AX\tminus{} & 8A+ \\
        Novel & & 8A+ \\
        \bottomrule
    \end{tabular}
\end{table}

\begin{table}[ht!]
    \centering
    \caption{Simulation parameters for Simulation Set 2}
    \label{tab:Exp2-LIrrev-params}
    \smallgraybox{\hspace{-1.5ex}
        \begin{tabular}{>{\raggedright\arraybackslash}m{.10\columnwidth} >{\raggedright\arraybackslash}m{.85\columnwidth}}
            \textbf{PKH} & $\alpha = 0.9$ \; $s = 0.2$ \; $\lambda = 1$ \; $\beta^+ = 0.5$ \; $\beta^- = 0.3$ \; $\gamma = 0.1$ \\
            \textbf{ME} & $\alpha = 0.9$ \; $\lambda = 1$ \; $\beta^+ = 0.5$ \; $\beta^- = 0.3$ \; $\gamma = 0.1$ \; $\theta^E = 0.8$ \; $\theta^I = 0.1$ \\
            \textbf{LPH} & $\alpha^M = 0.9$ \; $\alpha^H = 0.9$ \; $\lambda = 1$ \; $\beta^+ = 0.5$ \; $\beta^- = 0.3$ \; $\gamma = 0.1$ \; $\theta^E = 0.8$ \; $\theta^I = 0.1$ \\
            \textbf{MLAB} & $\alpha = 0.5$ \; $\lambda = 1$ \; $\beta^+ = 0.5$ \; $\beta^- = 0.3$
        \end{tabular}
    }
\end{table}

In this experiment, three groups and two phases are considered. In Phase 1, Group \textsf{Learned Irrelevance} is given partial reinforcement to a cue \textsf{X}, and to a compound \textsf{AX}. Group \textsf{CS-preexposure} receives exposure to the same cue and compound stimulus, but in extinction. Group \textsf{Novel} receives no treatment. In Phase 2, all groups are trained in simple conditioning \textsf{A+}.

\begin{figure}[ht!]
    \centering
\includegraphics[width=.75\linewidth]{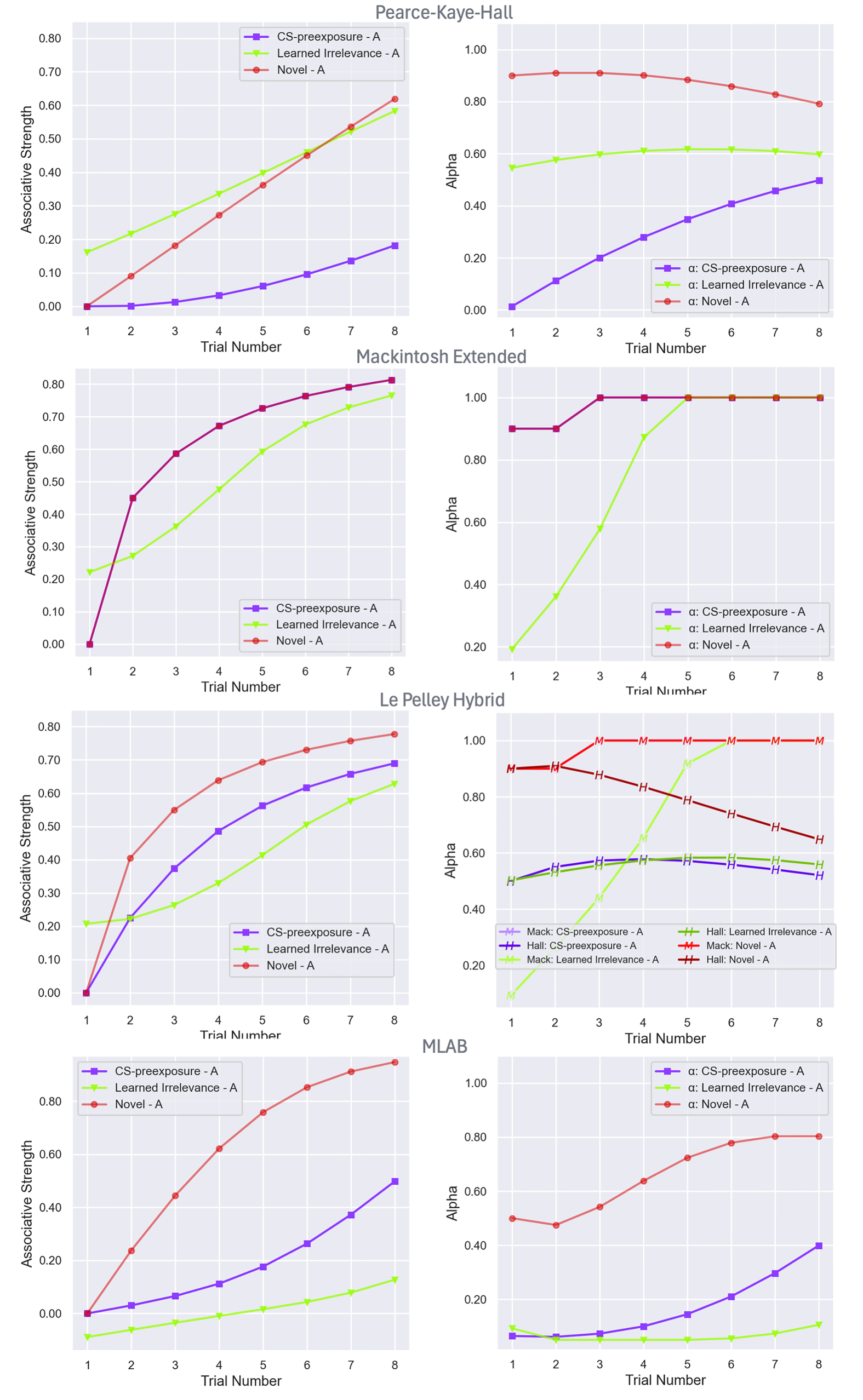}

\caption{Simulations of a simplified Learned Irrelevance \cite{lepelley2004}. The plots on the left show associative strength across 8 training trials during Phase 2 for the \textsf{Learned Irrelevance}, \textsf{CS-Preexposure}, and \textsf{Novel} groups. On the left side, the corresponding $\alpha$ value. From top to bottom, simulations of the Pearce, Kaye and Hall model, the Mackintosh Extended model, Le Pelley's Hybrid model and the MLAB model. Random trials included 500 sequences.}
\label{fig:Fig3-LePeley2004-LIrrel}
\end{figure}

\Cref{fig:Fig3-LePeley2004-LIrrel} displays the results of simulations of the different models considered, from top to bottom: Pearce-Kaye-Hall, Mackintosh Extended, Le Pelley's Hybrid, and the MLAB models. The left-hand side shows the associative strength acquired by \textsf{A} across the 8 learning trials, and, on the right, corresponding attentional $\alpha$ values.

Simulations of the PKH model predict a strong latent inhibition effect following exposure to \textsf{A} in Group \textsf{CS-preexposure}, as compared with the control Group \textsf{Novel}. That is, conditioning of \textsf{A} in the former group was substantially delayed in relation to the latter. Conversely, and against empirical evidence, the associative strength acquired by the same stimulus in Group \textsf{Learned Irrelevance} is stronger during early conditioning training. Phase 1 $\alpha$ values were omitted to save space, but the initial value during Phase 2 reveals the variation. Consistently, with the conditioning results, the cue's associability at the beginning of Phase 2 is lowest in Group \textsf{CS-preexposure}, intermediate in Group \textsf{Learned Irrelevance}, and highest in Group \textsf{Novel}.

Mackintosh Extended simulations, on the other hand, show a significant learned irrelevance effect, with cue \textsf{A} learning more slowly in Group \textsf{Learned Irrelevance} than in the other two, which do not differ. A consonant pattern is evident in the analysis of the cue's associability -- with an initial considerably lower associability of \textsf{A} in \textsf{Learned Irrelevance} than in the other groups.

Neither of these two models, Pearce-Kaye-Hall, Mackintosh Extended, seems able to replicate empirical observations, e.g.,\cite{bennett1995learned,baker1979preexposure}, according to which learning should proceed more slowly in Group \textsf{Learned Irrelevance} than in Group \textsf{CS-preexpusure}, which, in turn, should be retarded relative to learning in Group \textsf{Novel}.

LPH's and MLAB's simulations, however, are consistent with real data. Both models predicted that uncorrelated presentations of the CS and the US in Phase 1 would delay the formation of an association between them more than CS preexposure alone. Simulations from these models followed the same pattern, although they showed clear differences in magnitude. More significantly, their attentional analyses differ. In the former case, Le Pelley's simulations suggest that this retardation in conditioning is mainly due to the drastic reduction of the associability attention, $\alpha^M$, to \textsf{A} in Group \textsf{Learned Irrelevance} in comparison to that in Group \textsf{CS-preexposure}, during Phase 1. This distinct decline in learning rate results in a large difference at the start of Phase 2 training.

A different pattern of associability is predicted by the MLAB model. The associability values of cue \textsf{A} at the start of the conditioning phase are comparable, indicating a similar decrease in both groups during Phase 1, although somewhat more marked in Group \textsf{Learned Irrelevance}. This lower value is sustained throughout Phase 2 training, whereas the associability of \textsf{A} in Group \textsf{CS-preexposure} progressively increased during training, leading to faster conditioning. Consistent with the observed rate of conditioning, $\alpha$ in Group \textsf{Novel} also rose.

This set of simulations demonstrates the suitability of PALMS for directly comparing model predictions under the same design and equivalent conditions. More critically, visualisations of $\alpha$ values enable the researcher to immediately trace potential contributors to the observed results, offering valuable theoretical insights that could, in turn, assist in the formulation of hypotheses and guide new empirical research.

\clearpage{}
\subsection{Latent inhibition and reversed latent inhibition}
\label{haselgrove_etal_experiment}

Diverging from traditional CS processing views of latent inhibition \cite{schmajuk1996latent,E_PH_elaboration,Hall1991,pearce1980model,two_kinds_of_attention,Millard2022}, Byrom, Msetfi, and Murphy \cite{Byrom_2018} suggested that learning is delayed when the novelty of the target cue and the context in which conditioning takes place are homogeneous, and is facilitated when there is a mismatch between cue and context novelty. Recently, Haselgrove and collaborators \cite{novelty_mismatch_haselgrove} reported a set of experiments that further explored Byrom and collaborators' proposal using a sizeable number of stimuli to define contextual cues. Their research found a reverse latent inhibition effect, that is, a facilitation of learning with stimulus exposure in humans when the novelty/familiarity of the experimental context was manipulated. These results were in consonance with Byrom, Msetfi, and Murphy's ideas.

Although we do not anticipate that any of these models would be able to account for the reported effect, we nonetheless simulate the experiment to illustrate the implementation's potential to process a large number of CSs and provide insights into the theoretical discussion and aid further elaborations. Thus, in this paper, we attempted to replicate Haselgrove et al. \cite{novelty_mismatch_haselgrove} Experiment 2 and carried out simulations using the Pearce-Kaye-Hall and Le Pelley's Hybrid models.

Haselgrove et al.'s experiment consisted of two phases. In Phase 1, participants in Group Distractors-Novel received 20 trials of the preexposed stimulus and 60 distractors. In Group Distractors-Repeated, participants were exposed to 20 preexposed stimulus trials and to 15 trials of each of four distractors. During Phase 2, all participants received training with 20 pairings of the preexposed stimulus and a target stimulus (the outcome) and 20 pairings of a novel stimulus and the same outcome. Furthermore, in Group Distractors-Novel, 256 different distractors were interspaced in the training session, and 20 additional new distractors were presented, each paired with the target stimulus. In Group Distractors-Repeated, 64 trials of each of the distractors presented in the previous phase were presented, and an extra 5 of each of them were paired with the outcome. All trials were semi-randomly intermixed.

In the simulation presented here, the groups were labelled as \textsf{D-novel} and \textsf{D-repeated} for simplicity, the preexposed stimulus was labelled as \textsf{A}, the distractors as \textsf{D\tcaret{}1} to \textsf{D\tcaret{}60}, the novel conditioned stimulus and the distractors in Phase 2 as \textsf{B} and \textsf{S\tcaret{}1} to \textsf{S\tcaret{}256}, respectively, and the additional 20 distractors in \textsf{D-novel} Phase 2 paired to the outcome as \textsf{R\tcaret{}1} to \textsf{R\tcaret{}20}.

To our knowledge, this is the only published implementation that allows input and computation of a large number of stimuli, such as those described above. The design and parameter details for these simulations are displayed in \cref{tab:Exp3-ReversedLI,tab:Exp3-ReversedLI-params}.
\hfill 
\begin{table}[ht!]
    \centering
    \caption{Haselgrove et al. 2025 reversed latent inhibition design for Simulation Set 3}
    \label{tab:Exp3-ReversedLI}
    \sffamily
	\footnotesize
    \begin{tabular}{l l l}
        \toprule
        & Phase 1 & Phase 2 \\
        \midrule
		D-novel & 20A\tminus{}/D\tcaret{}1\tminus{}/ \dots{} /D\tcaret{}60\tminus{} & 20A+/20B+/\phantom{64}S\tcaret{}1\tminus{}/ \dots{} / S\tcaret{}256\tminus{}/ R\tcaret{}1+/ \dots{} / R\tcaret{}20+ \\
        D-repeated & 20A\tminus{}/D\tcaret{}1\tminus{}/ \dots{} /D\tcaret{}4\tminus{} & 20A+/20B+/64D\tcaret{}1\tminus{}/ \dots{} /64D\tcaret{}4\tminus{}/5D\tcaret{}1+/ \dots{} /5D\tcaret{}4+ \\
        \bottomrule
    \end{tabular}
\end{table}
\hfill \break
\begin{table}[ht!]
    \centering
    \caption{Simulation parameters for Simulation Set 3}
    \label{tab:Exp3-ReversedLI-params}
	\smallgraybox{\hspace{-1.5ex}
		\begin{tabular}{>{\raggedright\arraybackslash}m{.12\columnwidth} >{\raggedright\arraybackslash}m{.80\columnwidth}}
			\textbf{PKH} & $\alpha = 0.35$ \; $s = 0.2$ \; $\lambda = 0.8$ \; $\beta^+ = 0.5$ \; $\beta^- = 0.3$ \; $\gamma = 0.1$ \\
			\textbf{LPH} &  $\alpha^M = 0.2$ \; $\alpha^H = 0.05\ (0.9)$ \; $\lambda = 0.8$ \; $\beta^+ = 0.5$ \; $\beta^- = 0.3$ \; $\gamma = 0.1$ \; $\theta^E = 0.8$ \; $\theta^I = 0.1$
		\end{tabular}
	}
\end{table}

\begin{figure}[ht!]
    \centering
\includegraphics[width=.80\linewidth]{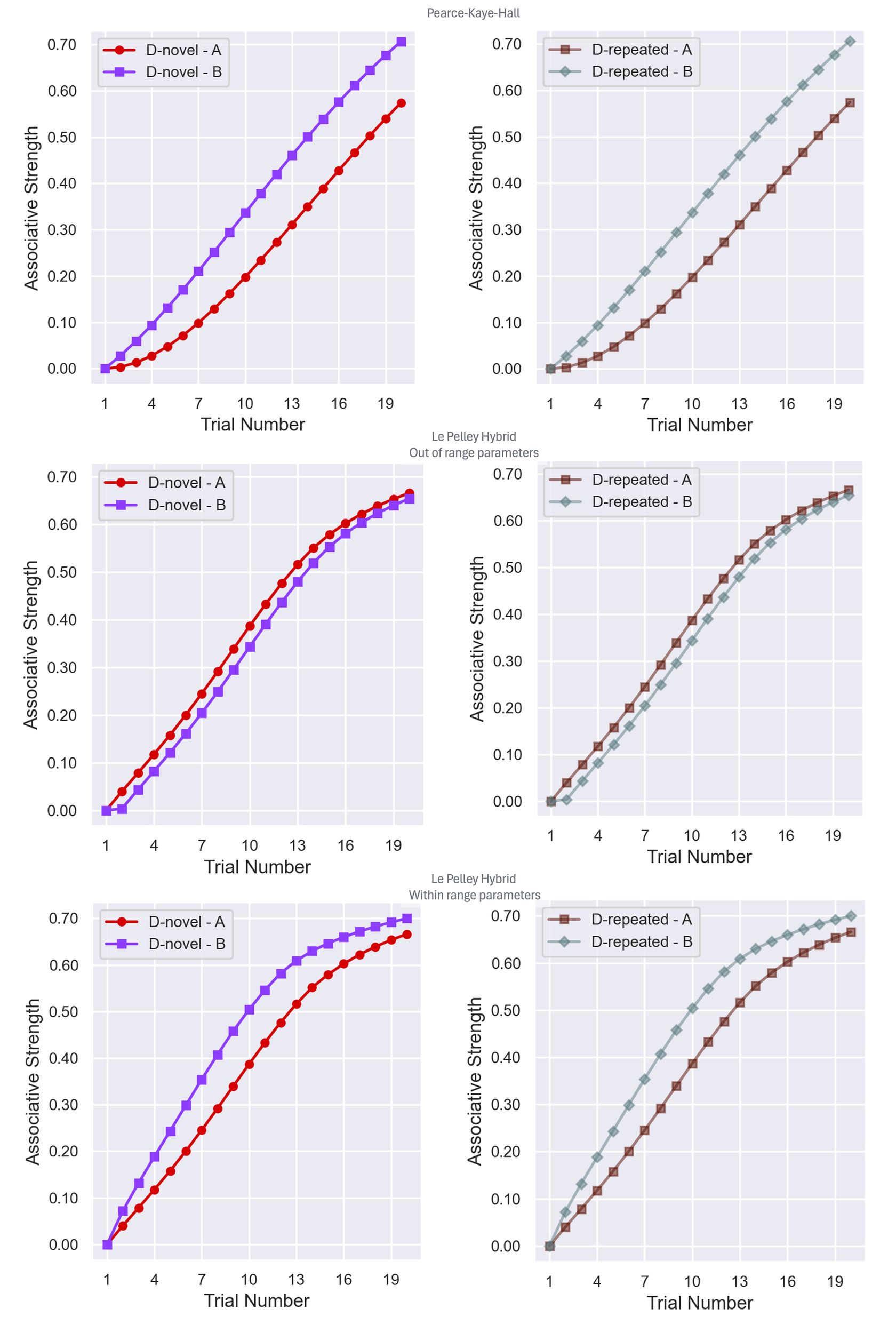}
\caption{A simulation of Haselgrove et al. 2025 Experiment 2 \cite{novelty_mismatch_haselgrove}. The top panel shows the results predicted by the Pearce-Kaye-Hall model during the test in Phase 2, for Group \textsf{D-novel} (left panel) and Group \textsf{D-repeated} (right panel). The middle panel displays corresponding results as simulated by Le Pelley's Hybrid model, using an out-of-range $\alpha^H$ value (0.05). On the bottom panel, Le Pelley's Hybrid model predictions are displayed with parameters within the model's range ($\alpha^H=0.9$). Random trials included 200 sequences.}
\label{fig:Fig4-Haselgrove2025}
\end{figure}

The simulated results during the test phase are shown in \cref{fig:Fig4-Haselgrove2025}. Simulations of the Pearce-Kaye-Hall model (top panel) unfold the predicted associative strength for the preexposed stimulus \textsf{A} and the control stimulus \textsf{B} in Group \textsf{D-novel} (on the left panel) and Group \textsf{D-repeated} (on the right) across training. A quick inspection of these results reveals an identical pattern across both preexposure conditions, consistent with a standard latent inhibition effect. That is, the acquisition of predictive value was delayed for the preexposed stimulus relative to the control in both groups. This pattern contradicts the experimental results, which showed facilitation of learning (reversed latent inhibition) in Group \textsf{D-novel} and a standard latent inhibition effect in Group \textsf{D-repeated}.

Although the differences are small, an opposite pattern to that predicted by Pearce-Kaye-Hall was observed when Le Pelley's Hybrid model of attention was used instead. Namely, reversed latent inhibition was predicted for both preexposed conditions when a set of very specific parameters was used.

By using an attentional salience, $\alpha^H$, value outside the range specified in the model (0.5-1, \cite{lepelley2004}, p. 227; Eq 24), Le Pelley's Hybrid is capable of predicting a reversed latent inhibition effect, that is, of anticipating a somewhat faster conditioning to the preexposed stimulus than conditioning to the control stimulus. However, contrary to the empirical results, Le Pelley's model makes the same prediction for both experimental conditions. Conversely, if $\alpha^H$, is kept within the theory-imposed boundaries, then, like the Pearce-Kaye-Hall model, Le Pelley's model predicts a speed of conditioning consistent with standard latent inhibition.

In the face of these results, we are compelled to conclude that neither of the two models is capable of replicating Haselgrove et al. results \cite{novelty_mismatch_haselgrove}.

However, researchers may capitalise on Le Pelley's model predictions and explore potential ways this approach, or others, could be adapted to fully account for at least some of the observations. That is, given the results, Le Pelley's potential to account for a reversed latent inhibition effect is evident. By manipulating the model's parameters outside their given range, the effect emerges. Strengthening this result would require, for instance, identifying new interactions between the postulated two $\alpha$ values, namely the model's attentional and salience associability, that may allow the model to predict learning facilitation under certain preexposure conditions.

More conceptual and formal research would nevertheless be needed to explain the differential effect generated by the distractors' repetition on the context-familiarity/novelty dimension. This outcome, and the cross-dimensional interaction reported by Haselgrove and collaborators \cite{novelty_mismatch_haselgrove} in the same paper, would remain unaccounted for and warrant still further theoretical research and potential new computational developments.

\clearpage{}
\subsection{Biconditional discriminations with compounds with different CS saliences}

The set of simulations below aims to showcase PALMS's ability to simulate experimental discriminations involving complex stimuli which may rely on configural learning.

Byrom and Murphy \cite{byrom2019cue} investigated the effect of within-compound overshadowing on complex stimulus discriminations, which are assumed to require some form of configural learning \cite{urcelay2025,mondragon2017associative} (but see \cite{delamater2017elemental}) when the component stimuli differed in salience. In Experiment 1, a group of human participants was presented with biconditional discriminations in which the physical salience of the stimuli relevant to solving the discrimination within a compound was either matched or mismatched. Another group of participants received a control uniconditional discrimination task with identical stimuli. Byrom and Murphy hypothesised that differences in the physical salience of stimuli forming the compound would weaken the development of configural learning, needed to solve a biconditional discrimination, and therefore interfere with the acquisition of a discrimination in the mismatched condition compared to the matched compounds. Simpler discriminations, such as uniconditional discriminations, which supposedly do not require configural learning, should not be affected. Thus, salience discrepancies between the elements of the compounds in an uniconditional discrimination should not affect their development.

We carried out RW simulations of Byrom and Murphy \cite{byrom2019cue} experiment and compared them to the Mackintosh Extended predictions. In both cases, we incorporated configural cues into the simulations. To the best of our understanding, this is the only publicly available computational implementation of the Mackintosh Extended model and the first time that configural cues are computed within the model framework.

\Cref{tab:Exp4-Bicond,tab:Exp4-Bicond-params} show the design of this experiment as entered in the interface. There were two independent groups. In Group \textsf{Bicond}, the task consisted of a biconditional discrimination, in which the relevant stimuli were either of matched or of mismatched salience. A control Group \textsf{Unicond} presented an uniconditional discrimination which also included a matched and a mismatched salience condition. Following the empirical set-up, stimuli \textsf{A}, \textsf{B}, \textsf{X} and \textsf{Y} were always high salience (0.5), whereas \textsf{R} and \textsf{S} were low salience (0.25). Letters \textsf{R} and \textsf{S} followed by a prime character (') were used to denote low salience stimuli irrelevant to discrimination. Letters \textsf{A}, \textsf{B}, \textsf{X} and \textsf{Y} followed by a caret 1 symbol (\tcaret{}1) were used to indicate high salience but irrelevant stimuli.

\begin{table}[ht!]
    \centering
    \caption{Byrom and Murphy, 2019 biconditional discrimination design for Simulation Set 4}
    \label{tab:Exp4-Bicond}
    \sffamily
    \scriptsize
    \begin{tabular}{l l l}
        \toprule
        & Matched & Mismatched \\
        \midrule
        \multirow{4}{*}{Bicond}
        & 12AXR'+/12AYR'-/ & 12AX\tcaret{}1R+/12AY\tcaret{}1R+/ \\
        & 12BXR'-/12BYR'+/ & 12BX\tcaret{}1R-/12BY\tcaret{}1R-/ \\
        & 12AXS'+/12AYS'-/ & 12AX\tcaret{}1S-/12AY\tcaret{}1S-/ \\
        & 12BXS'-/12BYS'+ & 12BX\tcaret{}1S+/12BY\tcaret{}1S+ \\
        \midrule
        \multirow{4}{*}{Unicond}
        & 12AX\tcaret{}1R'+/12AY\tcaret{}1R'+/ & 12A\tcaret{}1X\tcaret{}1R+/12A\tcaret{}1Y\tcaret{}1R+/ \\
        & 12BX\tcaret{}1R'-/12BY\tcaret{}1R'-/ & 12B\tcaret{}1X\tcaret{}1R+/12B\tcaret{}1Y\tcaret{}1R+/ \\
        & 12AX\tcaret{}1S'+/12AY\tcaret{}1S'+/ & 12A\tcaret{}1X\tcaret{}1S-/12A\tcaret{}1Y\tcaret{}1S-/ \\
        & 12BX\tcaret{}1S'-/12BY\tcaret{}1S'- & 12B\tcaret{}1X\tcaret{}1S-/12B\tcaret{}1Y\tcaret{}1S- \\
        \bottomrule
    \end{tabular}
\end{table}

\begin{table}[ht!]
    \centering
    \caption{Simulation parameters for Simulation Set 4}
    \label{tab:Exp4-Bicond-params}
    \smallgraybox{\hspace{-1.5ex}
        \begin{tabular}{>{\raggedright\arraybackslash}m{.10\columnwidth} >{\raggedright\arraybackslash}m{.85\columnwidth}}
            \textbf{RW} & $\alpha(\text{high}) = 0.5$ \; $\alpha(\text{low}) = 0.25$ \; $\lambda = 1$ \; $\beta^+ = 0.5$ \; $\beta^- = 0.3$ \\
            \textbf{ME} & $\alpha(\text{high}) = 0.5$ \; $\alpha(\text{low}) = 0.25$ \; $\lambda = 1$ \; $\beta^+ = 0.5$ \; $\beta^- = 0.3$ \; $\theta^E = 0.3$ \; $\theta^I = 0.1$ \\
            \textbf{} &  \rule{20em}{0.4pt} \\
            \textbf{} & $\alpha_{q(i,j,k),\text{matched}} = 0.05$ \; $\alpha_{q(i,j,k),\text{mismatch}} = 0.01$
        \end{tabular}
    }
\end{table}

\begin{figure}[ht!]
    \centering
\includegraphics[width=1\linewidth]{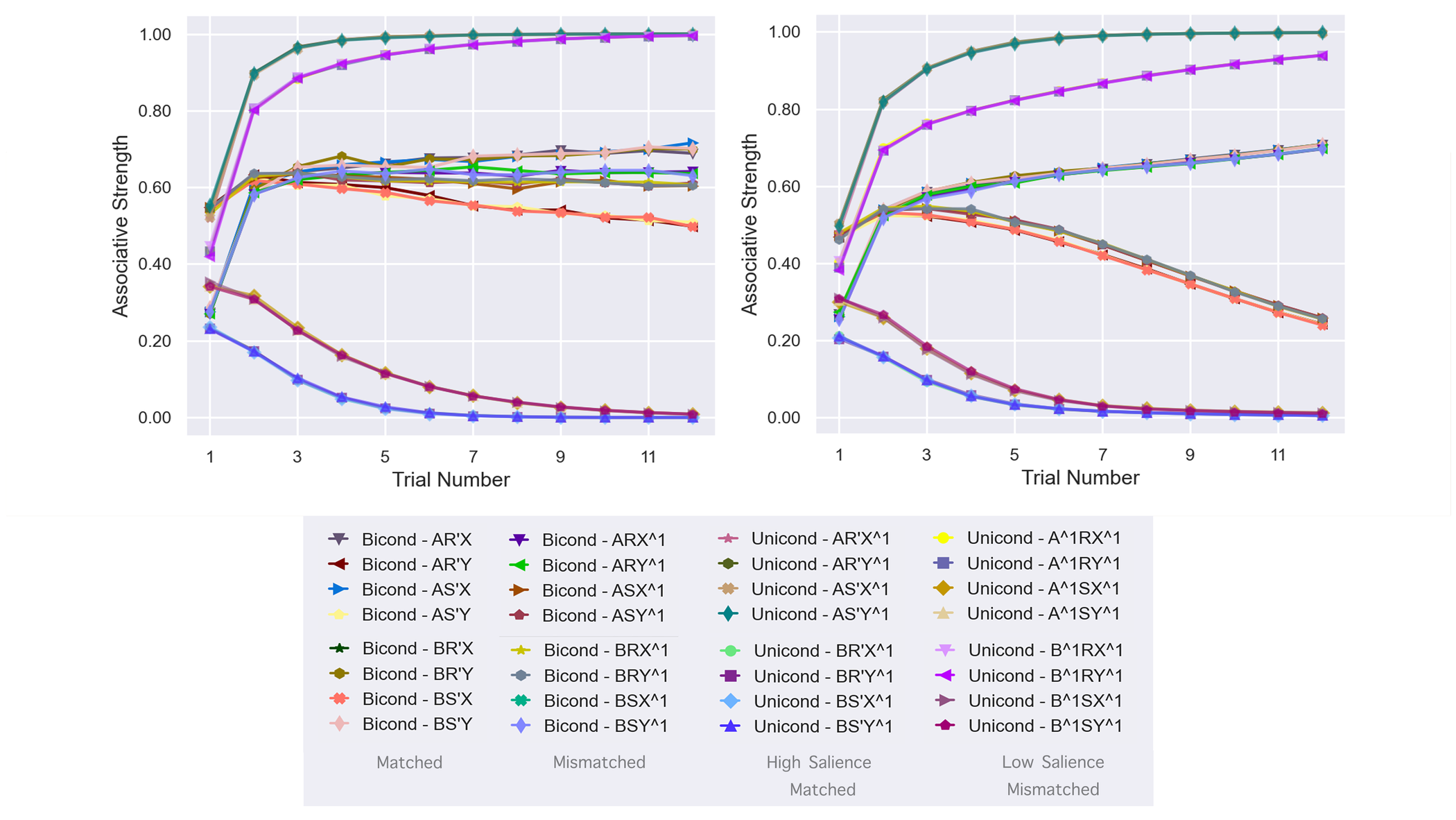}
\caption{Simulation of Byrom and Murphy 2019 Experiment 1 \cite{byrom2019cue} run with 1500 random sequences. The left panel shows the results predicted by the RW model with configural cues for Group \textsf{Bicond} (middle lines) and Group \textsf{Unicond} (top and bottom lines). The right panel displays Mackintosh Extended simulated associative strength for the same groups and conditions.}
\label{fig:Fig5-Byrom-Murphy2019}
\end{figure}

Consistent with the authors' prediction, when configural cues, as defined by Wagner and Rescorla \cite{r_w_inhibition_pavlovian}, are allowed to form and take part in the process of learning \cite{mondragon2013extension,RW_simulator,RW_simPLUS}, the RW model predicts that unequal stimulus salience will interfere with the acquisition of a biconditional discrimination (see \cref{fig:Fig5-Byrom-Murphy2019}). Compared to the uniconditional discriminations, which developed quickly (top and bottom lines), the biconditional discriminations were poorer (middle lines). In addition, as hypothesised, learning was impaired in compounds with mismatched salience compared with those with matched salience. However, because the simulation allowed configural cues to form in all conditions, learning was also hindered in compounds with mismatched salience in Group \textsf{Uncond}. Although configural cues are not needed to solve the discrimination, nothing prevents a system from taking advantage of them if formed. The general impairment due to mismatched salience, however, is conditional on assuming that the configural cues formed for compounds with unequal salience had a lower associability value than those resulting from compounds with equal salience stimuli. This is a reasonable assumption, since the rationale naturally follows the authors' posited rationale. If configural learning were indeed impaired in compounds with stimulus salience discrepancies, then we could assume that the effective salience of the resulting configural cues would be diminished as well.

On the contrary, under identical conditions, simulations with the Mackintosh extended model predict no differences due to mismatch salience in solving a biconditional discrimination, but surprisingly, it predicts that mismatch salience will hinder learning of an uniconditional discrimination. Similarly to the empirical results, the uniconditional discrimination developed faster and stronger than the biconditional one, but, unlike them, the models predict that unequal stimulus salience will interfere with learning a discrimination that allegedly does not require configural learning and can be solved by attending to individual stimuli.

Since both models, RW and Mackintosh Extended, incorporate a global error-correction term that enforces cue competition, the simulated results suggest that, in addition to cue competition, other factors may need to be considered to account for the observed empirical distinctive pattern. For example, in keeping with the postulates of the Mackintosh Extended model, the distribution of attention to cues will be affected by the inclusion of additional cues. Competition from configural cues may thus comparatively enhance attention to the relevant individual stimuli and may facilitate learning. The higher their initial associability (like in the matched salience condition), the stronger the facilitation.

\subsection{Hall-Pearce negative transfer}

We introduce this set of simulations to emphasise the implementation's potential to fine-tune discrimination between theoretical predictions, thereby helping refine experiments and models.

Hall and Pearce \cite{hallpearce1979} presented further evidence of a stimulus loss of associability resulting from repeated exposure. In this research, unlike in standard latent inhibition experiments \cite{latent_inhibition,Lubow_1995,kutlu2022dopamine,mondragon_2010}, the CS was presented paired with a weak US. Following this training, the target stimulus was paired with a strong US. Compared to a condition in which the target stimulus did not receive initial reinforced training, learning was delayed. Hall-Pearce negative transfer \cite{savastano1998temporal}, \cite {rodriguez2011reinforced}, \cite{negative_transfer} has been studied since then to explore the conditions that underlie changes in stimulus associability and has presented as evidence against Mackintosh's conceptualisation of predictive attention.

In their Experiment 2 \cite{hallpearce1979}, three groups of rats were trained with a Tone and a strong shock (\maltese) \medspace \normalsize during Phase 2. The groups differed on the Phase 1 treatment. Group Tone-shock received 66 trials of a tone followed by a weak shock (\small{+})\normalsize. In Group Light-shock, instead, a light served as the CS. Finally, Group T alone received 66 isolated presentations to the tone. For simplicity, we renamed the groups \textsf{T+}, \textsf{L+}, and \textsf{T\textminus}, respectively. We conducted simulations of this experiment with the Pearce-Kaye-Hall, Mackintosh Extended, Le Pelley's Hybrid, and MLAB models. The details of the design and parameters used can be found in \cref{tab:Exp5-NegTrans,tab:Exp5-NegTrans-params}.

\begin{table}[ht!]
    \centering
    \caption{Hall and Pearce, 1979 negative transfer design for Simulation 5}
    \label{tab:Exp5-NegTrans}
    \sffamily
    \footnotesize
    \begin{tabular}{l l l}
        \toprule
        & Phase 1 & Phase 2 \\
        \midrule
        T+ & 66T+ & 6T\maltese \\
        L+ & 66L+ & 6T\maltese \\
        T\tminus{} & 66T\tminus{} & 6T\maltese \\
        \bottomrule
    \end{tabular}
\end{table}

\begin{table}[ht!]
    \centering
    \caption{Simulation parameters for Simulation 5}
    \label{tab:Exp5-NegTrans-params}
    \smallgraybox{\hspace{-1.5ex}
        \begin{tabular}{>{\raggedright\arraybackslash}m{.12\columnwidth} >{\raggedright\arraybackslash}m{.80\columnwidth}}
            \textbf{PKH} & $\alpha = 0.5$ \; $\gamma = 0.1$ \; $\beta^- = 0.1$ \\
            \textbf{ME} & $\alpha = 0.5$ \; $\beta^- = 0.1$ \; $\gamma = 0.1$ \; $\theta^E = 0.8$ \; $\theta^I = 0.1$ \\
            \textbf{LPH} & $\alpha^M = 0.9$ \; $\alpha^H = 0.9$ \; $\beta^- = 0.1$ \; $\gamma = 0.1$ \; $\theta^E = 0.8$ \; $\theta^I = 0.1$ \\
            \textbf{MLAB} & $\alpha = 0.8$ \; $\beta^- = 0.1$ \; $\lambda^{\maltese} = 1$ \; $\lambda^{\scriptstyle{+}} = 0.25$ \; $\beta^{\,\maltese} = 0.8$ \; $\beta^{\,\scriptstyle{+}} = 0.05$
        \end{tabular}
    }
\end{table}

\begin{figure}[ht!]
    \centering
\includegraphics[width=.75\linewidth]{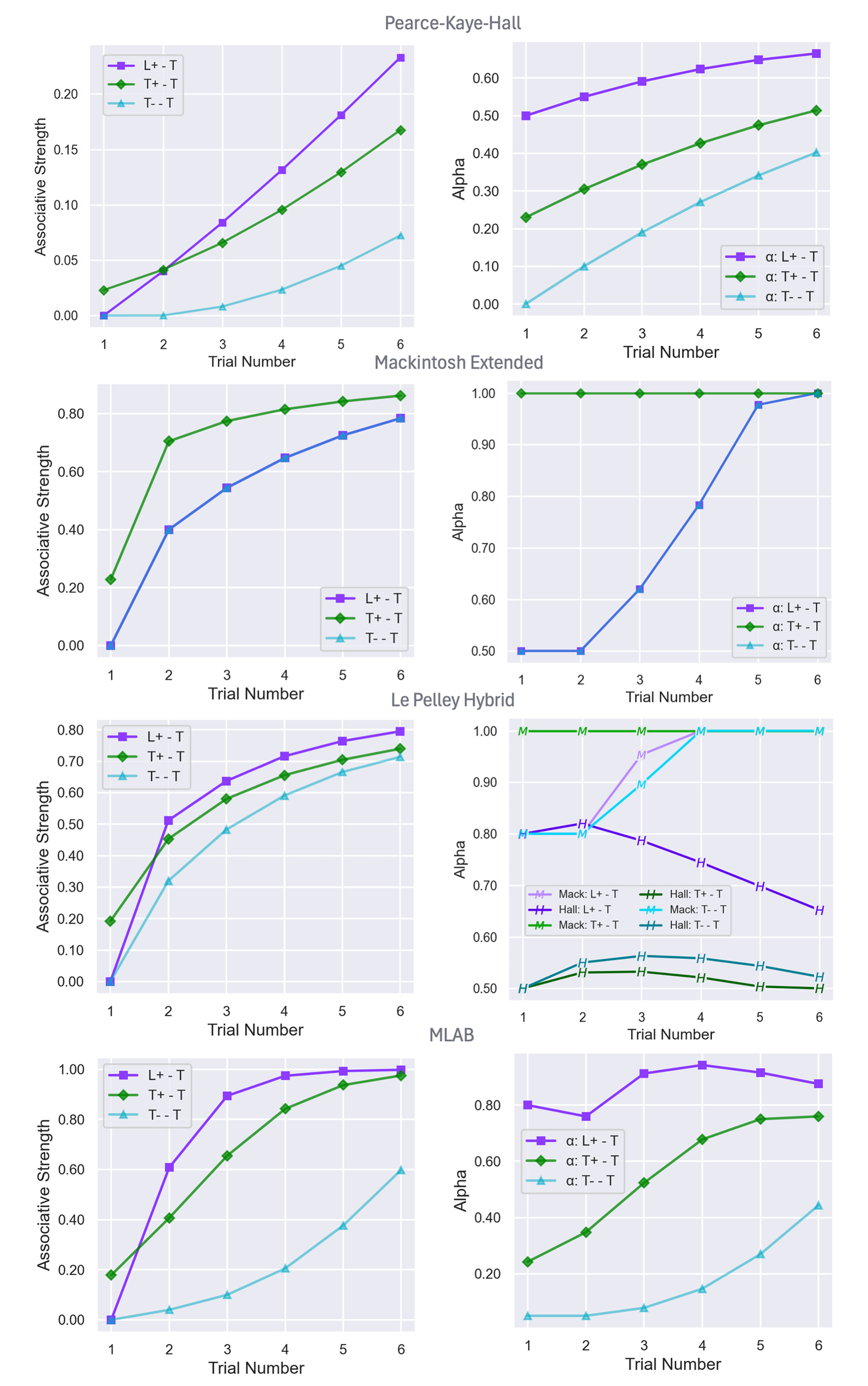}
\caption{Simulation of Hall and Pearce Experiment 2 \cite{hallpearce1979}. Associative strength predictions for each group, \textsf{T+}, \textsf{L+}, and \textsf{T\textminus}, during Phase 2 are displayed on the left panels. The corresponding $\alpha$ values on the right. Simulations from different models are shown top-to-bottom: Pearce-Kaye-Hall, Mackintosh Extended, Le Pelley's Hybrid and MLAB.}
\label{fig:Fig6-Hall-Pearce-1979}
\end{figure}

Simulation results for this experiment during Phase 2 are shown in \cref{fig:Fig6-Hall-Pearce-1979}. Unsurprisingly, a simulation carried out with the Mackintosh Extended model (top panel) could not replicate the empirical pattern of results. Contrary to Hall and Pearce's results, learning in Group \textsf{T+} developed faster than in the other two groups, which did not differ in their rates of learning. This pattern corresponds to the high discrepancy in $\alpha$ values (right panel) at the beginning of Phase 2 training. When the target stimulus was reinforced during Phase 1 in Group \textsf{T+}, its attentional rate increased from a default $\alpha=0.5$, value that remained invariable in the remaining groups, to a value around 1 at the start of Phase 2 conditioning.

Conversely, simulations of the Pearce-Kaye-Hall, Le Pelley's Hybrid and the MLAB models displayed a pattern of results consistent with the empirical data. All three models predicted faster learning in Group \textsf{L+}, for which the target CS was not preexposed, in comparison to Group \textsf{T-}, which received non-reinforced exposure to the tone. Critically, Group \textsf{T+}, in which the target stimulus was paired with a weak US in Phase 1, showed an intermediate speed of learning, replicating the Hall-Pearce Negative transfer effect. Values of the attentional parameter $\alpha$ at the start of Phase 2 can account for these differences in rate of learning. Simulations of PKH and MLAB indicate that the target stimulus $\alpha$ decayed during Phase 1 in Group \textsf{T-}, and remained high or even increased in value (MLAB model) in Group \textsf{T+}, as shown at the beginning of Phase 2 training, in comparison to the default value in Group \textsf{L+}. Le Pelley's $\alpha^H$ values in groups \textsf{T+} and \textsf{T-} drastically decreased during Phase 1 from the default (0.8) at the beginning of Phase 1 to 0.5 at the start of Phase 2, but did not differ at that point. The faster rate shown in Group \textsf{T+} relative to Group \textsf{T-} can be attributed to a higher $\alpha^M$ value at the start of conditioning. In Group \textsf{L+}, the interplay between a higher initial $\alpha^H$ and the rapid acceleration of $\alpha^M$ accounts for its faster learning.

The results of the different simulations are fully consistent with the traditional assumptions made from the models. They therefore serve to emphasise the validity and reliability of PALMS' simulations. They also highlight the value of the implementation in facilitating model comparison and analysis of results.
 
\section{Discussion}

Models of associative learning provide formal mechanisms to describe the theoretical principles underlying how humans and animals learn and adapt to their environment. Among them, attentional proposals have generated challenging new hypotheses and significantly advanced empirical research. However, associative learning models have customarily been expressed as non-instantiated mathematical descriptions, lacking well-specified dynamic methods. By focusing on static relations between variables, formal models do not, on their own, possess sufficient step-by-step conditions for computational implementation. This is a serious drawback, since computational models allow researchers to formulate accurate predictions. They provide a simplified and well-defined framework in which the body of knowledge can be assessed and tested in context, thus enabling qualitative and quantitative comparison of competing hypotheses and facilitating the formulation of new ones. 

Computational implementations of attentional mechanisms are necessary to enable researchers to generate precise and reliable simulations of empirical outcomes. It has been argued \cite{UsesAbuses} that simulations serve two main purposes, namely, to generate precise definitions required to implement a model, making it “accountable”, and to facilitate fast and accurate execution of calculations. Simulations are also an indispensable step in the cycle of theory formation and refinement. 

Although computational implementations of some of the associative learning models presented in the paper exist (e.g., \cite{mondragon2013extension,WRSimPaper,RW_simulator,RW_simPLUS,PH_simulator,matlab1}), their specifications are defined and coded independently and thus, devoid of a common testing environment that would allow for comparable model evaluation. In addition, most of them have become either obsolete, due to their outdated system requirements or because of the slow running time of the programming languages with which they were designed. Others have been designed to fulfil specific tasks, or they lack a comprehensive feature integration \cite{calmr_app}.

The simulations presented in this paper distinctly demonstrate that, despite the deceptive simplicity of these attentional models of learning, predictions are often conflated when multiple factors are involved. This observation highlights the need for suitable computational tools capable of producing fast, accurate simulations of the entire experimental setup.

We implemented a computational environment that provides a user-friendly, specialised tool for evaluating associative learning phenomena. PALMS includes features that enable the execution of complete experimental designs for five different learning models, with special emphasis on attentional approaches. The simulation capabilities include random trials, different parameters per stimulus, per-phase parameters, and compound stimulus calculation. It also empowers researchers to compute configural cues and configural-cue compounds for all models and to simulate experiments with hundreds of stimuli, features that we believe are unique to PALMS. The comprehensive integration of all characteristics enables immediate comparisons across different models and experimental setups, greatly improving research reproducibility and efficiency. This makes PALMS an effective tool for simulating existing associative learning experiments and testing novel design ideas.

In the first set of simulations, we showed that PALMS can be instrumental for isolating the contributions of different model features when some are shared across models.
The second set of simulations clearly illustrated that PALMS is a valuable tool for comparing model predictions seamlessly under the same design and conditions. Furthermore, visualisations of $\alpha$ values can help researchers immediately identify potential sources and intervening factors behind their results, assisting them in formulating and developing new hypotheses.

Next, we explored the potential of the current implementation to compute and graphically display what, in terms of learning theory, is a vast number of stimuli. Although neither of the currently implemented models was able to account for the interaction context novelty/familiarity found by Haselgrove et al. \cite{novelty_mismatch_haselgrove}, the results suggested the potential of Le Pelley's Hybrid model \cite{lepelley2004} to replicate their reversed latent inhibition effect.

The following set of simulations showcased PALMS's ability to simulate experimental discriminations involving complex stimuli and potentially configural learning. Among the novel features introduced in PALMS, these simulations highlighted the use of configural cues. To the best of our understanding, this is the only available algorithmic implementation that computes configural cues for PKH, ME and Le Pelley's models.
If we assume that configural learning is within the Mackintosh Extended framework \cite{lepelley2004}, understood as the integration of additional emergent cues into the computation, then the presence of new cues would entail a redistribution of attention. In this context, competition from configural cues may interfere with the learning of the nominal cues.

The final series of simulations outlined in this paper underscores the implementation's potential to help discriminate theoretical predictions, thereby contributing to the refinement of experiments and models. The results of the various simulations were entirely consistent with the conventional assumptions derived from the models, thereby reaffirming the validity and reliability of PALMS' simulations. Additionally, the selected simulations demonstrate the utility of the implementation for model comparison and result analysis.

Being open-source, PALMS and its authors welcome ongoing development and contributions from the scientific community. Future expansions may incorporate additional associative learning models. Another prospective development of the code involves integrating learning algorithms capable of computing qualitatively different outcomes. Such an extension would require defining the motivational direction and sensory properties of the outcomes, as well as formulating theoretical assumptions about the potential interference of different predictor-outcome associations during the learning process. These enhancements would expand the applicability and utility of the current implementation, thereby supporting advancements in both theoretical and empirical research within the field of associative learning. It is anticipated that this tool will be widely used by neuroscientists to support research and education.

We invite researchers to engage with PALMS, provide feedback, and contribute to the ongoing improvement and expansion of this collaborative tool via forks and pull requests on GitHub, as well as by contacting the authors directly.
 
\section{Acknowledgments}

We want to thank Abel Emanuel Bancu, and other students of the MSc Artificial Intelligence at City St George's, University of London, for testing and suggesting improvements to the PALMS program and Professor Eduardo Alonso for providing access to the computational and space resources, the Robin Milner Lab, needed to develop this project.
 
\nolinenumbers{}

\bibliography{bibliography}

\newpage 

\begin{appendices}

\section{Pseudocode}

\label{S1_Appendix}
\textbf{Pseudocode for implemented learning models and phase-running procedures.}
\DontPrintSemicolon

\SetKwProg{Switch}{switch}{}{end}
\SetKwProg{Case}{case}{}{}
\SetKwProg{Default}{default}{}

\SetKwProg{Function}{Function}{}{end}

\SetKw{FeIn}{in}
\SetKwFunction{Sum}{sum}
\SetKwFunction{Avg}{avg}
\SetKwFunction{Add}{add}

\textbf{Models of associative learning.}
\label{rw_algorithm_subsection}
\begin{algorithm}[H]
\label{rw_algorithm}
\caption{Step function for the Rescorla-Wagner model}
\label{RWalg}
\Function{Step}{
  \KwIn{Stimulus, RunParameters}
  \KwOut{Stimulus}
  \BlankLine

  $\left< V, V^E, V^I, \alpha, \alpha^M, \alpha^H, S \right> \gets \text{Stimulus}$\;
  $\left< \beta, \lambda, \mathit{sign}, \Sigma, \Sigma^E, \Sigma^I \right> \gets \text{RunParameters}$\;
  \BlankLine

  $V \gets V + \alpha \cdot \delta_V$\;
  \Return{Stimulus}
}
\end{algorithm}

\begin{algorithm}[H]
\caption{Step function for the Pearce-Kaye-Hall model}
\label{PKHalg}
\Function{Step}{
  \KwIn{Stimulus, RunParameters}
  \KwOut{Stimulus}
  \BlankLine

  $\left< V, V^E, V^I, \alpha, \alpha^M, \alpha^H, S \right> \gets \text{Stimulus}$\;
  $\left< \beta, \lambda, \mathit{sign}, \Sigma, \Sigma^E, \Sigma^I \right> \gets \text{RunParameters}$\;
  \BlankLine

  $\rho \gets \lambda - (\Sigma^{E} - \Sigma^{I})$\;
  \BlankLine

  \If{$\rho \ge 0$}{
    $V^{E} \gets V^{E} + \beta \cdot \alpha \cdot \lambda \cdot S$\;
  }
  \Else{
    $V^{I} \gets V^{I} + \beta^{-} \cdot \alpha \cdot | \rho | \cdot S$\;
  }
  \BlankLine

  $\alpha \gets \gamma \cdot | \rho | + (1 - \gamma) \cdot \alpha$\;
  $V \gets V^{E} - V^{I}$\;
  \Return{Stimulus}
}
\end{algorithm}

\begin{algorithm}[H]
\caption{Step function for Mackintosh Extended model}
\label{MEalg}
\Function{Step}{
  \KwIn{Stimulus, RunParameters}
  \KwOut{Stimulus}
  \BlankLine

  $\left< V, V^E, V^I, \alpha, \alpha^M, \alpha^H, S \right> \gets \text{Stimulus}$\;
  $\left< \beta, \lambda, \mathit{sign}, \Sigma, \Sigma^{E}, \Sigma^{I} \right> \gets \text{RunParameters}$\;
  \BlankLine

  $\rho \gets \lambda - (\Sigma^{E} - \Sigma^{I})$\;
  $V^{E}_{X} \gets \Sigma^{E} - V^{E}$,\quad $V^{I}_{X} \gets \Sigma^{I} - V^{I}$\;
  \BlankLine

  \If{$\rho > 0$}{
    $\Delta V^{E} \gets \alpha \cdot \beta_{p} \cdot ( 1 - V^{E} + V^{I} ) \cdot | \rho |$\;
    $\alpha \gets \alpha - \theta^{E} \cdot ( \left| \lambda - V^{E} + V^{I} \right|
                                 - \left| \lambda - V^{E}_{X} + V^{I}_{X} \right| )$\;
    $V^{E} \gets V^{E} + \Delta V^{E}$\;
  }
  \ElseIf{$\rho < 0$}{
    $\Delta V^{I} \gets \alpha \cdot \beta^{-} \cdot ( 1 - V^{I} + V^{E} ) \cdot | \rho |$\;
    $\alpha \gets \alpha - \theta^{I} \cdot ( \left| | \rho | - V^{I} + V^{E} \right| - \left| | \rho | - V^{I}_{X} + V^{E}_{X} \right| )$\;
    $V^{I} \gets V^{I} + \Delta V^{I}$\;
  }
  \BlankLine

  \tcp{Clamp $\alpha$ to minimum and maximum values.}
  $\alpha \gets \min(\max(\alpha, 0.05), 1)$\;
  $V \gets V^{E} - V^{I}$\;
  \Return{Stimulus}
}
\end{algorithm}

\begin{algorithm}[H]
\caption{Step function for the Le~Pelley Hybrid model}
\label{LPHalg}
\Function{Step}{
	\KwIn{Stimulus, RunParameters}
	\KwOut{Stimulus}
	\BlankLine

	$\left< V, V^E, V^I, \alpha, \alpha^M, \alpha^H, S \right> \gets \text{Stimulus}$\;
	$\left< \beta, \lambda, \mathit{sign}, \Sigma, \Sigma^E, \Sigma^I \right> \gets \text{RunParameters}$\;
	\BlankLine

	$\rho \gets \lambda - (\Sigma^E - \Sigma^I)$\;
	$V^E_X \gets \Sigma^E - V^E$,\quad $V^I_X \gets \Sigma^I - V^I$\;
	\BlankLine

	$\Delta V^E \gets 0,\ \Delta V^I \gets 0$\;
	\If{$\rho \ge 0$}{
		$\Delta V^E \gets \alpha^H \beta_p \cdot \alpha^M ( 1 - V^E + V^I ) | \rho |$\;
		$\alpha^H \gets \alpha^H - \theta^E \alpha^M
		( | \lambda - V^E + V^I | - | \lambda - V^E_X + V^I_X | )$\;
	}
	\Else{
		$\Delta V^I \gets \alpha^H \beta^- \cdot \alpha^M \cdot (1 - V^I + V^E) \cdot | \rho |$\;
		$\alpha^H \gets \alpha^H - \theta^I \cdot \alpha^M
		( \left| | \rho | - V^I + V^E \right| - \left| | \rho | - V^I_X + V^E_X \right| )$\;
	}
	\BlankLine

	$\alpha^M \gets \gamma \cdot | \rho | + (1 - \gamma) \cdot \alpha^M$\;
	\BlankLine

	\tcp{Clamp $\alpha^H$ and $\alpha^M$ to minimum and maximum values.}
	$\alpha^H \gets \min(\max(\alpha^H, 0.05), 1)$,\quad
	$\alpha^M \gets \min(\max(\alpha^M, 0.5), 1)$\;
	\BlankLine

	$V^E \gets V^E + \Delta V^E$,\quad
	$V^I \gets V^I + \Delta V^I$\;
	$V \gets V^E - V^I$\;

	\Return{Stimulus}
	}
\end{algorithm}

\begin{algorithm}[H]
\caption{Step function for the MLAB model}
\label{ARWalg}

\Function{Step}{
  \KwIn{Stimulus, RunParameters}
  \KwOut{Stimulus}
  \BlankLine

  $\left< V, V^E, V^I, \alpha, \alpha^M, \alpha^H, S \right> \gets \text{Stimulus}$\;
  $\left< \beta, \lambda, \mathit{sign}, \Sigma, \Sigma^E, \Sigma^I \right> \gets \text{RunParameters}$\;
  \BlankLine

  $\alpha \gets \left| \lambda - \Sigma \right|$\;
  $V \gets V + S \cdot \alpha \cdot \left| \lambda \right|$\;
  \Return{Stimulus}
}
\end{algorithm}

\textbf{Running individual phases.}

\begin{algorithm}[H]
	\caption{Sequential Phase Running}
	\label{sequential_pseudocode}

	\Function{SequentialPhase}{
		\KwIn{Model, Trials, Parameters, Stimuli, per-phase~$\beta$, per-phase~$\lambda$}
		\KwOut{hist, Stimuli}
		\BlankLine

		$\left< \alpha, \alpha^M, \alpha^H, S, \lambda, \beta^+, \beta^-, \gamma, \theta^E, \theta^i \right> \gets \mathrm{Parameters}$\;
		\If{per-phase~$\beta$ is present}{
			$\beta^+ \gets \text{ per-phase } \beta$\;
		}
		\If{per-phase~$\lambda$ is present}{
			$\lambda \gets \text{ per-phase } \lambda$\;
		}
		\BlankLine

		\ForEach{compounds, US \FeIn Trials}{
			\Switch{US}{
				\Case{\texttt{``++''}}{
					$\beta, \mathit{sign} \gets 2 \cdot \beta^+, 1$\;
				}
				\Case{\texttt{``+''}}{
					$\beta, \mathit{sign} \gets \beta^+, 1$\;
				}
				\Case{\texttt{``\textminus{}''}}{
					$\beta, \lambda, \mathit{sign} \gets \beta^-, 0, -1$\;
				}
			}
			\BlankLine

			$\Sigma\;\;\, \gets \Sum(Stimuli.V\;\;\,)$\;
			$\Sigma^E     \gets \Sum(Stimuli.V^E)    $\;
			$\Sigma^I\, \gets \Sum(Stimuli.V^I\;) $\;
			$\mathrm{run\_parameters } \gets \left< \beta, \lambda, \mathit{sign}, \Sigma, \Sigma^E, \Sigma^I \right>$\;
			\BlankLine

			hist[compounds].\Add(Stimuli[compounds].V)\;
			\ForEach{cs \FeIn compounds}{
				$\left< V, V^E, V^I, \alpha, \alpha^M, \alpha^H, S \right> \gets \text{Stimuli}[\text{cs}]$\;
				hist[cs].\Add(V[cs])\;
				\BlankLine

				\tcp{\texttt{Model.run} modifies the parameters of the stimulus \texttt{cs}}
				Model.step(Stimuli[cs], $\mathrm{run\_parameters}$)\;
			}
		}
		\BlankLine

		\Return{hist, Stimuli}
	}
\end{algorithm}

\begin{algorithm}[H]
	\caption{Randomised Phase Running}
	\label{random_pseudocode}

	\Function{RandomisedPhase}{
		\KwIn{\textnumero{}, Model, Trials, Parameters, Stimuli, per-phase~$\beta$, per-phase~$\lambda$}
		\KwOut{hist, Stimuli}
		\BlankLine

		trial\_stimuli \textleftarrow{} []\;
		trial\_hist \textleftarrow{} []\;
		\For{trial \textleftarrow{} 1 \KwTo \textnumero{}}{
			Shuffle(Trials)\;
			\BlankLine

			hist, final\_stimuli \textleftarrow{} SequentialPhase(Model, Trials, Parameters, Stimuli, per-phase~$\beta$, per-phase~$\lambda$)\;
			trial\_hist.\Add(hist)\;
			trial\_stimuli.\Add(final\_stimuli)\;
		}
		\BlankLine

		\tcp{Return the average of each element of the stimulus history, and set the values of each CS to the average of the final value each trial.}
		hist \textleftarrow{} \Avg(trial\_hist)\;
		Stimuli \textleftarrow{} \Avg(trial\_stimuli)\;

		\Return{hist, Stimuli}
	}
\end{algorithm}

\clearpage{}
\section{CLI Help}

\label{S2_Appendix}
\textbf{Command-line help and interface reference.}

This is the result of the \texttt{-{}-help} command on the PALMS CLI, which is described in the main article.

\begin{lstlisting}[caption={\texttt{./PALMS cli -{}-help}, part 1/2}, float=htb]
usage: PALMS.py cli [-h] [--savefig filename] [--print-results]
                    [--save-results filename] [--singular-legend]
                    [--show-title] [--dpi DPI] [--output-width OUTPUT_WIDTH]
                    [--plot-phase phase_num] [--plot-experiments [group ...]]
                    [--plot-stimuli [conditioned_stimulus ...]]
                    [--plot-alpha | --no-plot-alpha]
                    [--plot-macknhall | --no-plot-macknhall]
                    [--plot-alphas | --no-plot-alphas]
                    [--part-stimuli | --no-part-stimuli]
                    [--adaptive-type {Rescorla Wagner,Pearce Kaye Hall,Mackintosh Extended,Le Pelley's Hybrid,MLAB Model}]
                    [--alpha α] [--alpha-mack αᴹ] [--alpha-hall αᴴ]
                    [--beta β⁺] [--beta-neg β⁻] [--lamda λ] [--gamma γ]
                    [--thetaE θᴱ] [--thetaI θᴵ] [--salience S]
                    [--habituation h] [--xi-hall ξ] [--num-trials №]
                    [--configural-cues | --no-configural-cues] [--rho ρ]
                    [--nu ν] [--kay κ] [--max-workers MAX_WORKERS]
                    [experiment_file]

positional arguments:
  experiment_file       Path to the experiment file.
\end{lstlisting}

\begin{lstlisting}[caption={\texttt{./PALMS cli -{}-help}, part 2/2}, float=htb]
options:
  -h, --help            show this help message and exit

Output parameters:
  --savefig filename    Instead of showing figures, one image per phase will be saved with the name "filename_1.png" ... "filename_n.png".
  --print-results       Instead of showing the plot, print the results of the experiment.
  --save-results filename
                        Instead of showing the plot, save the results of the experiment to a file.
  --singular-legend     Hide legend in output, and generate a separate image with just the legend. If run with --savefig, save it under "filename_legend.png".
  --show-title          Show title and phases to saved output.
  --dpi DPI             Dots per inch.
  --output-width OUTPUT_WIDTH
                        Width of the output

Plotting parameters:
  --plot-phase phase_num
                        Plot a single phase
  --plot-experiments [group ...]
                        List of experiments to plot.
  --plot-stimuli [conditioned_stimulus ...]
                        List of stimuli, compound and simple, to plot.
  --plot-alpha, --no-plot-alpha
                        Whether to plot the total alpha.
  --plot-macknhall, --no-plot-macknhall
                        Whether to plot the alpha Mack and alpha Hall.
  --plot-alphas, --no-plot-alphas
                        Whether to plot all the alphas, including total alpha, alpha Mack, and alpha Hall.
  --part-stimuli, --no-part-stimuli
                        Whether to plot part stimuli with US in addition to the regular plot.

Experiment Parameters:
  --adaptive-type {Rescorla Wagner,Pearce Kaye Hall,Mackintosh Extended,Le Pelley's Hybrid,MLAB Model}
                        Type of adaptive attention mode to use
  --alpha α             Alpha for all other stimuli
  --alpha-mack αᴹ       Alpha_mack for all other stimuli
  --alpha-hall αᴴ       Alpha_hall for all other stimuli
  --beta β⁺             Associativity of the US +.
  --beta-neg β⁻         Associativity of the absence of US +. Equal to beta by default.
  --lamda λ             Asymptote of learning.
  --gamma γ             Weighting how much you rely on past experinces on DualV adaptive type.
  --thetaE θᴱ           Theta for excitatory phenomena in Le Pelley blocking
  --thetaI θᴵ           Theta for inhibitory phenomena in Le Pelley blocking
  --salience S          Salience for all parameters without an individually defined salience. This is used in the Pearce & Hall model.
  --habituation h       Habituation delay for all parameters in the hybrid model.
  --xi-hall ξ           Xi parameter for Hall alpha calculation
  --num-trials №        Amount of trials done in randomised phases
  --configural-cues, --no-configural-cues
                        Whether to use configural cues
  --rho ρ
  --nu ν
  --kay κ
  --max-workers MAX_WORKERS
                        Maximum number of multiprocessing cores used in randomised phases. This is constrained by the total CPU count and number of trials.

  --alpha-[A-Z] α	Associative strength of CS A..Z.
  --alpha_mack-[A-Z] α
			Associative strength (Mackintosh) of CS A..Z.
  --alpha_hall-[A-Z] α
			Associative strength (Hall) of CS A..Z.
  --saliences-[A-Z] S
			Salience of CS A..Z.
  --habituations-[A-Z] h
			Habituation of CS A..Z.
\end{lstlisting}

Additionally, some options are available in the GUI.
These are useful for troubleshooting and controlling the maximum amount of multiprocessing workers running on random trials.

\begin{lstlisting}[caption=\texttt{./PALMS -{}-help}, float=htb]
usage: PALMS [-h] {cli,gui} ...

positional arguments:
  {cli,gui}
    cli       Run PALMS command-line interface. PALMS.py cli --help for mode information.
    gui       Run PALMS GUI interface. This is the default if no mode is selected.

options:
  -h, --help  show this help message and exit

usage: PALMS gui [-h] [--dpi DPI] [--fontsize FONTSIZE] [--fontscale FONTSCALE] [--screenshot-ready] [--debug] [--smoke-test] [--verbose]
                           [--max-workers MAX_WORKERS] [--spawn]
                           [initial_file]

positional arguments:
  initial_file          File to load initially

options:
  -h, --help            show this help message and exit
  --dpi DPI             DPI for shown and outputted figures.
  --fontsize FONTSIZE   Fontsize of the GUI; screenshots are taken in fontsize 16.
  --fontscale FONTSCALE
                        Scale of the font (overriden by --fontsize).
  --screenshot-ready    Hide guide numbers for easier screenshots.
  --debug               Whether to go to a debugging console if there is an exception
  --smoke-test          Run a smoke test: open the app, log everything, wait 5 seconds, close the app.
  --verbose, -v         Verbose logging.
  --max-workers MAX_WORKERS
                        Maximum number of multiprocessing cores used in randomised phases. This is constrained by the total CPU count and number of trials.
  --spawn               Force spawn instead of fork for multiprocessing. This should only have an effect on Linux, and is used for debugging.
\end{lstlisting}

\end{appendices}

\end{document}